\documentclass[letterpaper, 10 pt, journal, twoside]{IEEEtran}
\ifCLASSINFOpdf
\else
\fi
\usepackage{graphicx}
\usepackage{float}
\usepackage{subfigure}
\usepackage[justification=centering]{caption}
\usepackage{cite}
\usepackage{amsmath}

\usepackage[noend]{algpseudocode}
\usepackage{algorithmicx,algorithm}

\usepackage{array}

\usepackage{color}

\usepackage{makecell}
\usepackage{multirow}

\usepackage{url}

\usepackage{stfloats}

\usepackage{hyperref}

\begin{document}
%
\title{
	A Semi-automatic Oriental Ink Painting Framework for Robotic Drawing from 3D Models
}
%
%
%

\author{Hao Jin$^{1}$, Minghui Lian$^{1}$, Shicheng Qiu$^{1}$, Xuxu Han$^{1}$, Xizhi Zhao$^{1}$, Long Yang$^{1}$, \\ Zhiyi Zhang$^{1}$, Haoran Xie$^{2}$, Kouichi Konno$^{3}$ and Shaojun Hu$^{1}$%
\thanks{Manuscript received: March, 29, 2023; First revised April, 19, 2023; Second revised July, 8, 2023; Accepted August, 8, 2023.}
\thanks{This paper was recommended for publication by Editor Gentiane Venture upon evaluation of the Associate Editor and Reviewers' comments. This work was supported in part by the Natural Science Basis Research (NSBR) Plan of Shaanxi under Grant 2022JM-363, the Key Project of Shaanxi Provision-City Linkage under Grant 2022GD-TSLD-53 and the National Natural Science Foundation of China under Grant 61303124. (Corresponding author: Shaojun Hu.)} 
\thanks{$^{1}$College of Information Engineering, Northwest A\&F University, Yangling, Xianyang, China {\tt\footnotesize hsj@nwsuaf.edu.cn}}%
\thanks{$^{2}$Japan Advanced Institute of Science and Technology, Ishikawa, Japan {\tt\footnotesize xie@jaist.ac.jp}}%
\thanks{$^{3}$Faculty of Science and Engineering, Iwate University, Morioka, Japan {\tt\footnotesize konno@cis.iwate-u.ac.jp}}%
\thanks{Digital Object Identifier (DOI): see top of this page.}
}
%
%

\markboth{IEEE Robotics and Automation Letters. Preprint Version. Accepted August, 2023}
{Jin \MakeLowercase{\textit{et al.}}: A Semi-automatic Oriental Ink Painting Framework for Robotic Drawing from 3D Models} 

%



\maketitle

\begin{abstract}
Creating visually pleasing stylized ink paintings from 3D models is a challenge in robotic manipulation. We propose a semi-automatic framework that can extract expressive strokes from 3D models and draw them in oriental ink painting styles by using a robotic arm. The framework consists of a simulation stage and a robotic drawing stage. In the simulation stage, geometrical contours were automatically extracted from a certain viewpoint and a neural network was employed to create simplified contours. Then, expressive digital strokes were generated after interactive editing according to user's aesthetic understanding. In the robotic drawing stage, an optimization method was presented for drawing smooth and physically consistent strokes to the digital strokes, and two oriental ink painting styles termed as \textit{Noutan} (shade) and \textit{Kasure} (scratchiness) were applied to the strokes by robotic control of a brush's translation, dipping and scraping. Unlike existing methods that concentrate on generating paintings from 2D images, our framework has the advantage of rendering stylized ink paintings from 3D models by using a consumer-grade robotic arm. We evaluate the proposed framework by taking 3 standard models and a user-defined model as examples. The results show that our framework is able to draw visually pleasing oriental ink paintings with expressive strokes.
\end{abstract}


%
\IEEEpeerreviewmaketitle

\section{Introduction}
%
%
%
%
\IEEEPARstart{O}{riental} ink painting, also known as \textit{Shuimohua} in China and \textit{Suibokuga} in Japan, is an ancient monochrome painting art that abstracts complex object into a few expressive strokes and draws the strokes on a rice paper by skillfully using black ink, water and a soft brush. The philosophy of oriental ink painting is similar to the concept of ``\textit{less is more}" from the minimalist \cite{Chen2020}. However, it is difficult for common people or robots to draw aesthetically pleasing paintings because even a single stroke can produce considerable variations in shading, and an artist may spend years to master the drawing skills. Moreover, robotic painting from three-dimensional (3D) models involves several interdisciplinary fields such as robotics, computer graphics, human-computer interaction and art.

\begin{figure}[ht]
	\centering
	\includegraphics[width=\linewidth]{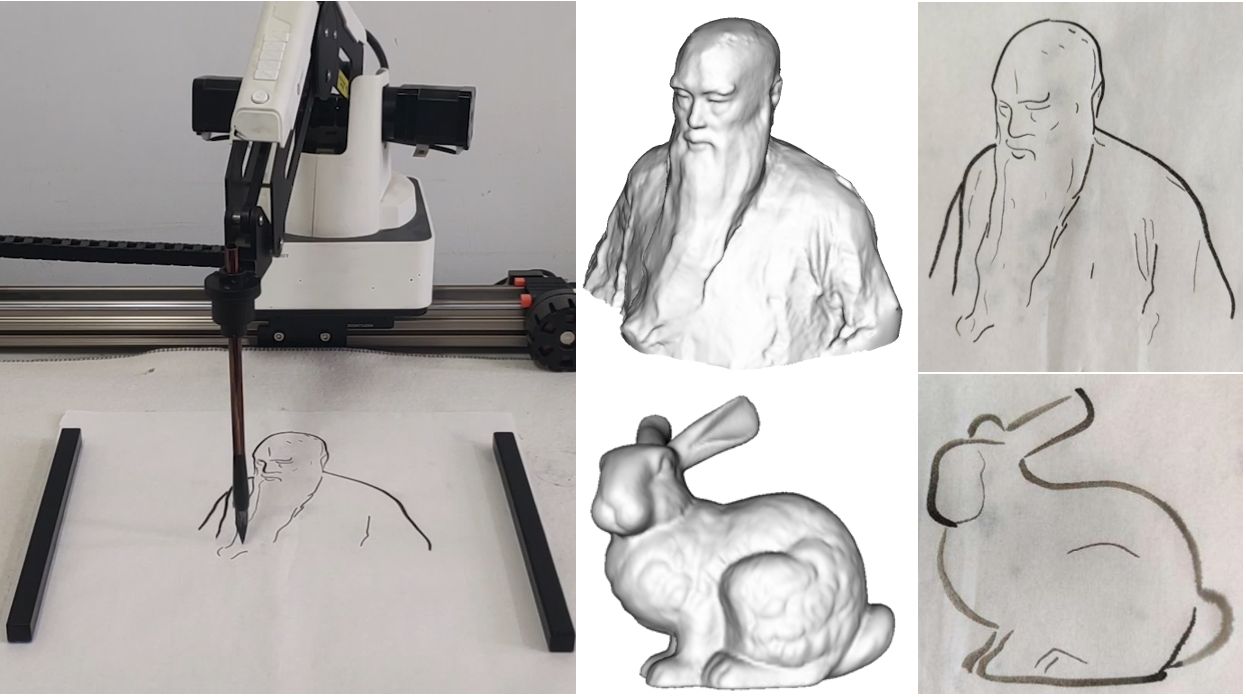}
	\captionsetup{justification=raggedright}
	\caption{Our framework can draw visually pleasing oriental ink painting given input 3D models. Left: the robotic arm drawing a portrait from a 3D model in ink painting style; Middle: input 3D models ``Youren'' sculpture and ``Stanford Bunny''; Right: robotic drawing results.}
	\label{fig:teaser}
\end{figure}

With the rapid advancement of Non-Photorealistic Rendering (NPR), a variety of digital arts have been created from images \cite{chen2017stylebank, tang2017animated}. However, most of these studies focus on generating digital works or imitate master's painting styles from two-dimensional (2D) data. Nowadays, it is convenient to reconstruct 3D models of real-world objects through depth sensors or multi-view photos \cite{Berger2017}. In comparison with 2D sketches and images, 3D models can better deliver the geometrical structure of objects. Thus, Grabli et al. \cite{Grabli2010} and Liu et al.\cite{Liu2021strokes} have studied the creation of stylized line drawings from 3D models. While these methods work well for generating digital paintings, it is a challenging task to realize physical paintings using a robotic arm, since the motion control of the robot and the interaction process among the brush, ink, and a paper are totally different from the digital rendering. Recently, Lindemeier et al. \cite{Lindemeier2015}, Scalera et al. \cite{Scalera2019}, and L\"{o}w et al. \cite{Low2022} have developed painting robots to achieve impressive oil painting, watercolor painting and portrait drawing from 2D images. However, the current robotic drawing methods haven't considered generating stylized ink paintings from 3D models. In this work, we design an oriental ink painting framework to physically draw stylized strokes from 3D models as shown in Fig. \ref{fig:teaser}. The main contributions of this work include: 

\begin{itemize}
	\item a practical user interface to vectorize and extract expressive digital strokes from 3D models by referring to both the simplified contours and the original models.
	\item an optimization and mapping method for drawing smooth and physically consistent strokes from the digital strokes.
	\item a realization of two oriental ink painting styles which are termed as \textit{Noutan} and \textit{Kasure} to create aesthetic effects of shading and scratchiness on a rice paper.
	\item a novel framework to convert 3D models to oriental ink paintings through interactive editing and robotic drawing.
\end{itemize}
%

\section{RELATED WORK}
The publications that are directly related to our work can be categorized into two groups: stylized line drawing methods from 3D models in simulation stage, and robotic drawing methods in real execution stage.

\subsection{Line Drawings from 3D Models} The problem of extracting and drawing expressive lines from 3D models is equivalent to the problem of how artists create line drawings from real objects, which is one of the challenging task in NPR. The detailed survey of line drawing methods from 3D models can be found in B\'{e}nard and Hertzmann's tutorial \cite{Benard2019}. Saito and Takahashi \cite{Saito1990} first created stylized contour lines and curved hatching from 3D models by using 2D image processing operations. Winkenbach and Salesin \cite{Winkenbach1994} extended the work of \cite{Saito1990} to generate more realistic line drawings from 3D models by integrating 2D and 3D rendering. Zhang et al. \cite{Zhang1999} presented a cellular automaton model to simulate the Suibokuga-like painting of 3D trees. DeCarlo et al. \cite{DeCarlo2003} proposed Suggestive Contours to improve the quality of line drawings by connecting perception knowledge and differential geometry. Grabli et al. \cite{Grabli2010} created vivid stylized line drawings such as Japanese big brush from 3D models based on the work of \cite{DeCarlo2003}. Judd et al. \cite{Judd2007} defined view-dependent curvature to generate Apparent Ridges that captured more detailed features than the Suggestive Contours. Recently, Uchida and Saito \cite{Uchida2020} introduced two fully convolutional neural networks to determine the intensity of strokes for stylized line drawings. Liu et al. \cite{Liu2021strokes} generated impressive stylized strokes for line drawings based on a combination of a differentiable geometric module and an image translation network. Although these methods have ability to extract feature lines and generate visually pleasing digital strokes from 3D models, they are unsuitable for creating physical stylized strokes for robotic drawing because the extracted lines are likely to be redundant or oversimplified, and the styles of digital strokes which are usually implemented by texture mapping are difficult to be realized in physical drawing. Inspired by the work of interactive inking for cleaning rough sketches \cite{Simo2018}, our work does not pursue fully automatic realization of line drawings but focuses on extracting a few non-overlapping expressive lines with a combination of existing automatic contour extraction method and human-computer interaction process.

\subsection{Robotic Drawing} The earliest work of robotic drawing could date back to Harold Cohen's exhibitions at the Computer Museum in 1995 \cite{Cohen1995}. Cohen developed a robotic drawing system ``AARON"  that could color its own paintings using a variety of brushes. Thanks to advances of sensor and AI technologies, the diversity, perception and collaboration skills of robots improved significantly. Calinon et al. \cite{Calinon2005} designed a $4$ Degrees of Freedom (DOFs) robotic arm to draw a portrait detected from a webcam using traditional image processing and inverse kinematics methods. L\"{o}w et al. \cite{Low2022} designed a robotic system ``drozBot" to draw artistic portraits from images based on a novel ergodic control algorithm. However, the above-mentioned methods can only create monochrome paintings. Recently, Lindemeier et al. \cite{Lindemeier2015} proposed a painting robot to realize colored oil painting styles from 2D images. Luo and Liu \cite{Luo2018} realized an interesting colored Cartoon style portrait painting using NPR techniques. Scalera et al. Karimov et al. \cite{karimov2023} developed a data-driven model for accurately color mixing and reproduced famous artworks based on a robotic arm. More recently, the deep learning methods have been developed to solve more complicated robotic drawing problems. Zhang et al. \cite{Zhang2019} trained a network to identify the type of individual strokes for intelligent calligraphy beautification. Gao et al. \cite{Gao2020} developed a robotic system for efficiently drawing portraits from images based on a combination network of Neural Style Transfer and Generative Adversarial Network. Bidgoli et al. \cite{Bidgoli2020} used deep learning method to learn stylized brushstroke from human artists and reproduce them through robotic painting. Furthermore, the surface of robotic drawing has been extended from 2D to 3D. Song et al. \cite{Song2018} presented an impedance control method that can draw user's 2D drawing on a 3D surface without vision support, and extended the drawing on a large and nonplanar surface with a mobile platform \cite{Song2023}. In contrast, Liu et al. \cite{Liu2022} developed a robotic system to draw 2D strokes on 3D surface based on scanned point clouds and a robust motion planning algorithm. While all these robotic drawing methods have varying focuses such as artistic portrait drawing \cite{Calinon2005, Luo2018, Gao2020, Low2022} or stylized colored painting from 2D images \cite{Lindemeier2015, Luo2018, Scalera2019, Schaldenbrand2023} or realistic calligraphy drawing \cite{Zhang2019} or robust drawing on 3D surfaces \cite{Song2018, Song2023, Liu2022}, they haven't considered creating stylized ink paintings from 3D models which requires understanding the abstraction technique from 3D models and the motion control of a robotic arm and a brush to generate the styles of oriental ink painting such as \textit{Noutan} and \textit{Kasure}.


\section{OVERVIEW}
\begin{figure}[h]
	\centering
	\includegraphics[width=\linewidth]{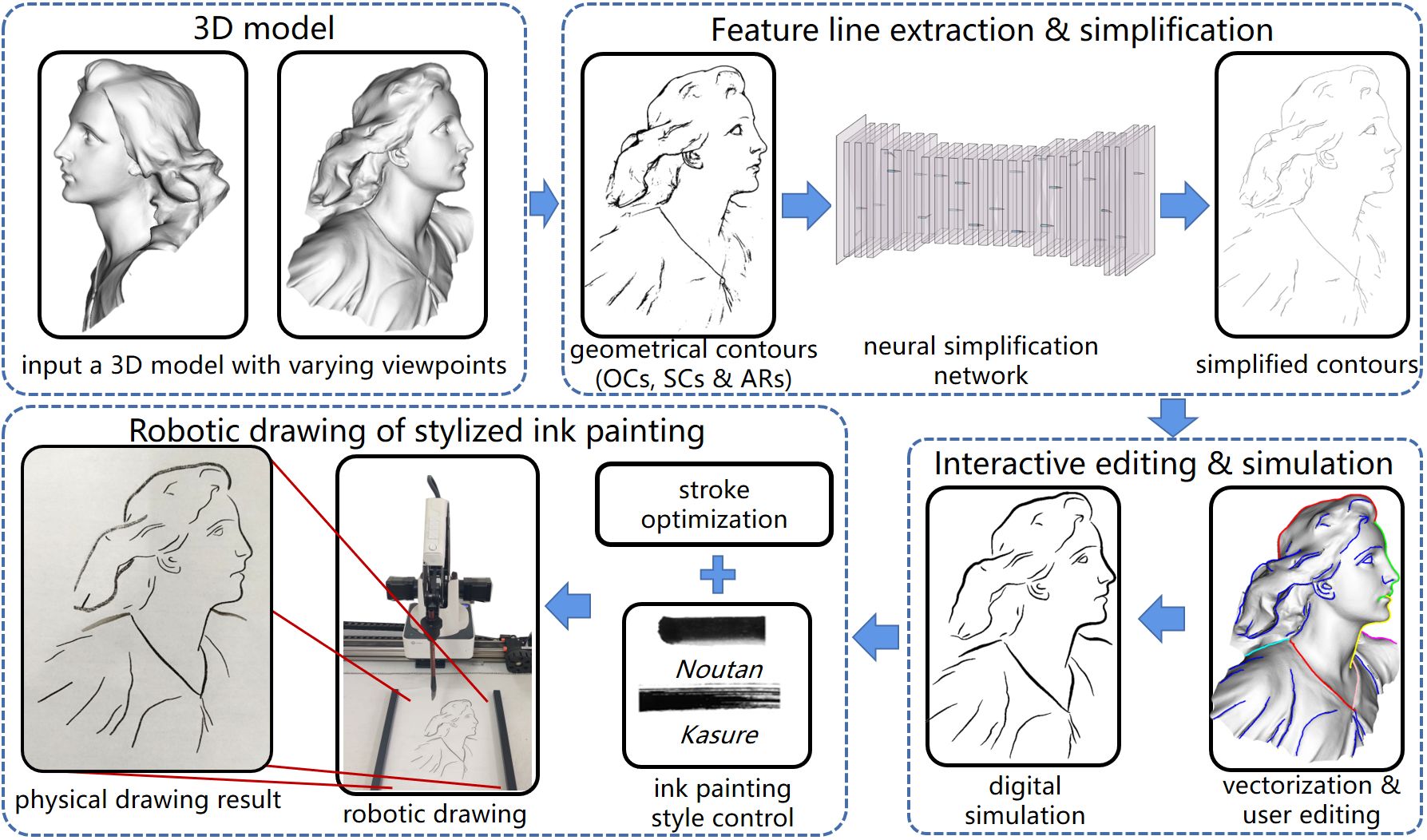}
	\captionsetup{justification=raggedright}
	\caption{Overview of our robotic drawing framework.}
	\label{fig: overview}
\end{figure}

In this work, we propose a semi-automatic robotic drawing framework to convert 3D models to physical stylized ink paintings. Fig. \ref{fig: overview} illustrates the workflow of our robotic drawing framework. We start off with a shaded 3D model with different viewpoints. To depict shapes by few featured strokes for ink painting, we extract geometrical contours from the 3D model and then employ a neural network to simplify the contours. Next, the simplified contour image is converted to a vector image by taking account of local and global curvatures, and a user interface is designed to allow users to pick, merge and insert expressive polylines in accordance with personal artistic perception for detailed enrichment. After user interaction, a digital ink painting with selected strokes is automatically generated in real-time. In order to control and draw physical strokes that are consistent to the digital strokes by using a robotic arm, we optimize the stoke trajectory and map the stroke properties from simulation space to physical space. Finally, two typical oriental ink painting styles which are termed as \textit{Noutan} and \textit{Kasure} are realized to add aesthetic effects of shading and scratchiness to the physical strokes.

\section{FEATURE LINE EXTRACTION AND SIMPLIFICATION} \label{sec: feature_extract}
In contrast to western paintings which usually concentrate on the realism of visual objects since the Renaissance \cite{Bao2016}, most oriental ink paintings emphasize the inner spiritual essence instead of exact imitation of objects. A typical technique is drawing few well-organized stylized strokes on a rice paper to leave white space for stimulating imagination. In order to depict shapes by few feature strokes, we extract geometry-based contours from 3D models and then utilize a neural network to simplify the contour lines for ink painting.

\begin{figure}[h]
	\centering
	\includegraphics[width=\linewidth]{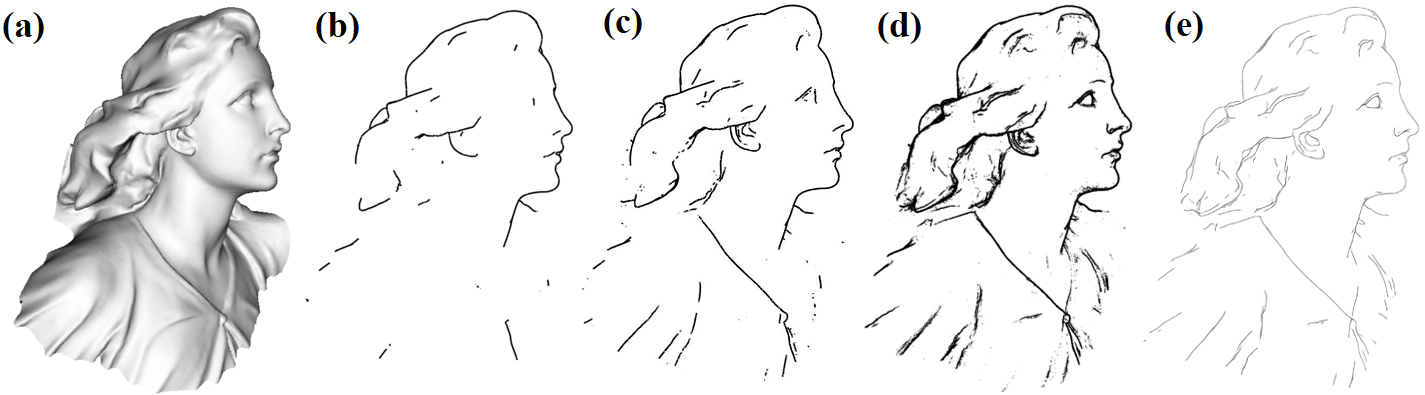}
	\captionsetup{justification=raggedright}
	\caption{(a) A rendered ``Lucy'' model from a given viewpoint and the corresponding (b) OCs, (c) OCs+SCs, (d) OCs+SCs+ARs, and (e) simplified contours.}
	\label{fig: lucy_contours}
\end{figure}

\subsection{Feature Line Extraction from 3D Models}
Inspired by the investigation that digital line drawings can depict 3D model that even match the artist's drawings \cite{Cole2009}, we select three typical geometry-based digital contours, namely occluding contours (OCs), suggestive contours (SCs)\cite{DeCarlo2003}, and apparent ridges (ARs) \cite{Judd2007}, as the basis for the creation of ink paintings from 3D models. Given a camera viewpoint and a 3D model, the combination of these contours produces a raster contour image as shown in Fig. \ref{fig: lucy_contours} (a-d), where OCs conveys the rough shape of the object, SCs and ARs add considerable amount of details to the 3D model.

\subsection{Neural Simplification Network}

\begin{figure}[h]
	\centering
	\includegraphics[width=\linewidth]{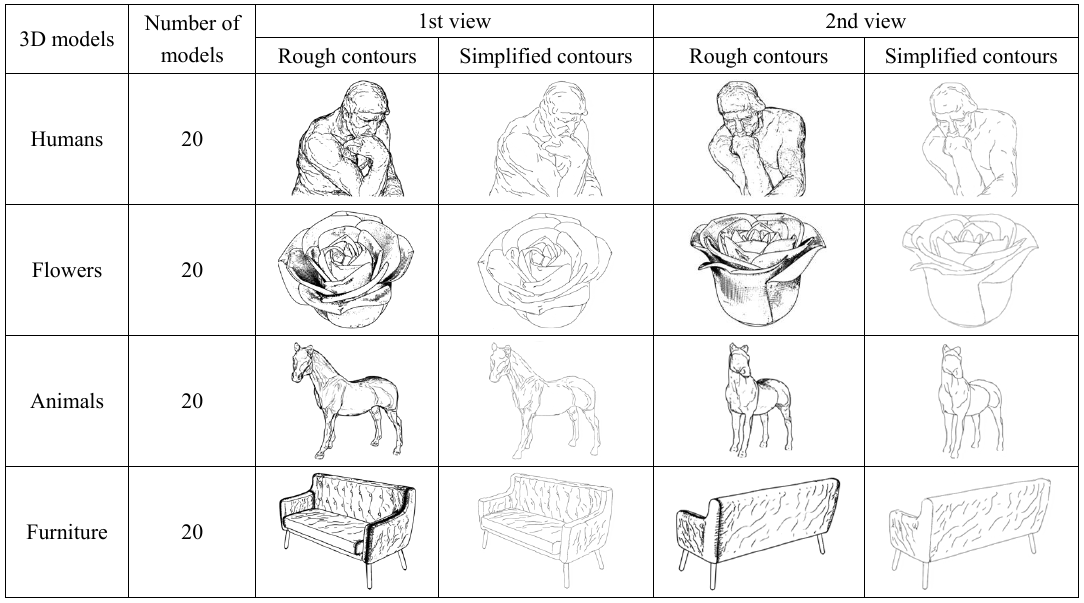}
	\captionsetup{justification=raggedright}
	\caption{ Dataset from two different views of 3D models for neural simplification network.}
	\label{fig: two_viewpoints_contours}
\end{figure}

Although the extracted contours are a reasonable description of 3D shape, it creates large amount of dense and detailed lines that are inapplicable for oriental ink paintings which focus on sparse stylized strokes. Motivated by the CNN-based neural network for sketch simplification \cite{Simo2016}, we employ the similar network with 3 down-convolution layers, 17 flat-convolution layers and 3 up-convolution layers to simplify the contour image. The loss function is defined as the $l^2$-norm of the difference between the output image and the target image. However, the dataset of the previous network is for cleaning rough pencil sketches. Therefore, we construct our contour simplification dataset through manual annotation.

\begin{figure}[htbp]
	\centering
	\includegraphics[width=\linewidth]{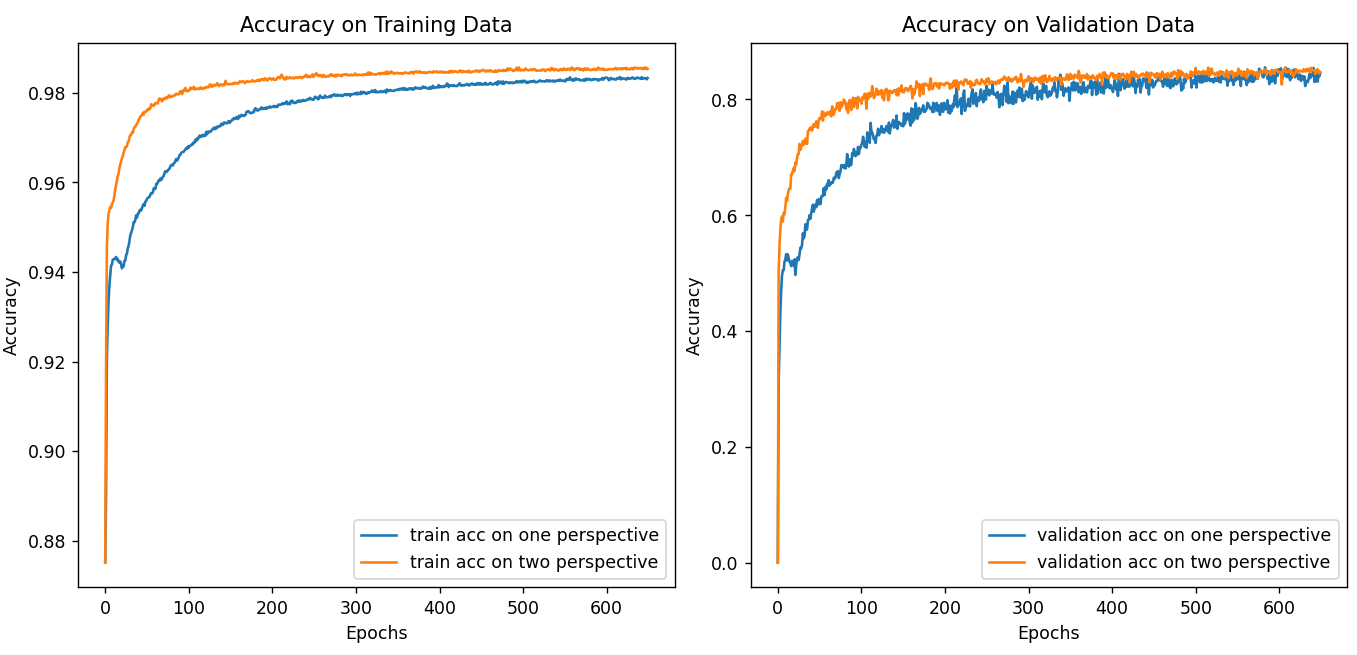}
	\captionsetup{justification=raggedright}
	\caption{Training and validation of neural simplification network.}
	\label{fig: new_train2}
\end{figure}

The dataset consists of two parts, one part is rough contours automatically extracted from 3D models and the other part is the corresponding feature lines selected by users. The 3D models in the dataset are constructed from $4$ categories including humans, flowers, animals and furniture, and each category has $20$ 3D model as shown in Fig. \ref{fig: two_viewpoints_contours}. To make the training robust, we augmented the dataset by applying image blurring, flipping and adding noise. Then, we measured precision by defining the contour pixels as ``True" and the rest pixels as ``False". The dataset was divided into training set, validation set and test set with a ratio of $8:1:1$. After $600$ epochs of iterative training with batch size of $8$, the precision of the network converged to $0.982$ and $0.853$ on the training set and the validation set when considering only one viewpoint, and the corresponding accuracy converged to $0.986$ and $0.857$ when considering two viewpoints as shown in Fig. \ref{fig: new_train2}, which indicated that increasing the number of viewpoints did not change the network accuracy much.

\section{VECTORIZATION AND INTERACTIVE EDITING} \label{sec: user_interface}
Thus far the simplified contours are represented by a raster image, it is a routine to convert the raster image to a vector image for ink painting. However, the vectorized image may possibly generate small disconnected or noisy contours. Moreover, the contours might be over-simplified that some expressive features are ignored. Therefore, we developed a user interface to interactively edit the simplified contours.

\subsection{Vectorization}
To generate smooth individual curves without junctions for robotic drawing, locating proper corners is critical because corners have significant variations in curvature. Thus, we use a robust corner detector to remove junctions or intersection points at which curves meet based on local and global curvatures \cite{He2008}. The local curvature of the $i$th contour pixel $(x_{i}, y_{i})$ in a raster image is denoted by $\kappa_{i}$. The pixels with the maxima of absolute curvature $|\kappa_{i}|$ are selected as corner candidates. However, round corners which are important for forming smooth curves might be incorrectly marked as candidates. Therefore, a global curvature threshold $\overline{\kappa_{i}}$ is defined to remove round corners by considering the mean curvature within a region of support $[L_{left}, L_{right}]$:
\begin{equation}\label{eqn: local_curvature}
	\begin{aligned}
		\overline{\kappa_{i}} = \frac{\mu}{L_{left}+L_{right}+1}\sum\limits_{j=i-L_{left}}^{i+L_{right}}|\kappa_{j}|
	\end{aligned}
\end{equation}
where $L_{left}$ and $L_{right}$ are the nearest curvature minima on both sides of pixel $i$, and $\mu$ is the splitting ratio to control the number of vectorized curves. If $|\kappa_{i}|$ of a pixel is less than $\overline{\kappa_{i}}$, it is classified as round corner and then should be removed from the corner candidates. Finally, the rest corner pixels are utilized as endpoints $X_{k}$ to form individual polylines, and each polyline is represented by a series of line segments $\left\lbrace{(X_{1}, X_{2}), (X_{2}, X_{3}), ..., (X_{k}, X_{k+1}), ..., (X_{n-1}, X_{n})}\right\rbrace$.

\begin{figure}[h]
	\centering
	\includegraphics[width=\linewidth]{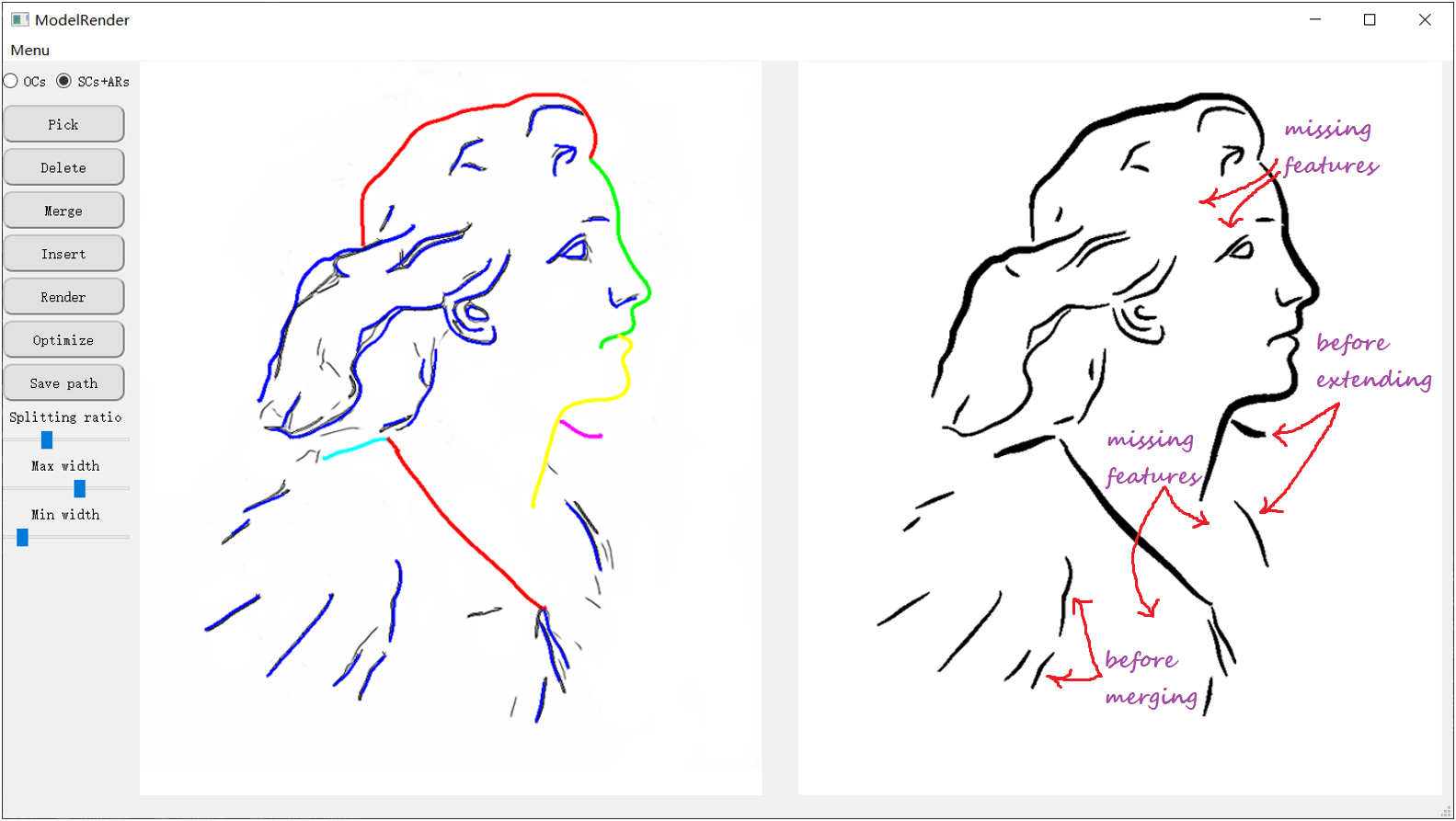}
	\text{(a) Interactive editing of vectorized contours.}
	\includegraphics[width=\linewidth]{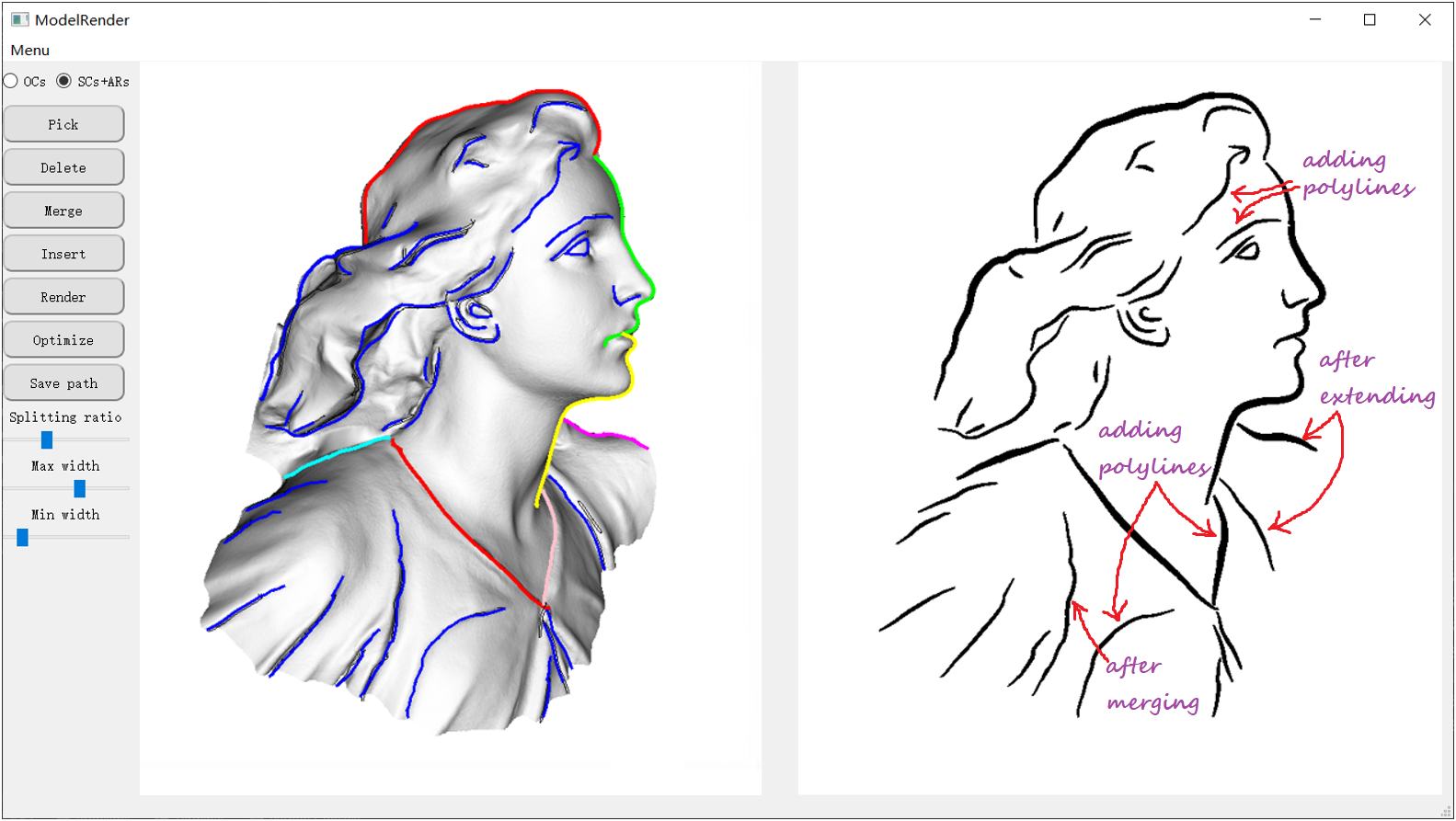}
	\text{(b) Refined result by referring to the 3D model.}
	\captionsetup{justification=raggedright}
	\caption{Human-computer interaction for extracting expressive strokes. (a) From the simplified contours, the ``Split'', ``Pick'' and ``Delete'' tools can be used to select outside contours (colored polylines except blue ones) and inside contours (blue polylines). (b) By referring to the original 3D model, we can refine the result of (a) by inserting, merging and extending polylines.}
	\label{fig: user_interface}
\end{figure}

\subsection{Interactive Editing}
To generate appropriate stylized strokes for oriental ink painting which emphasizes sparse and expressive contours, we developed a user interface (UI) as shown in Fig. \ref{fig: user_interface} to allow users to flexibly edit the simplified contours. The left view of the UI is designed for human-computer interaction, and the right view simultaneously shows the completed digital ink painting. The UI consists of several tools including ``Split'', ``Pick'', ``Delete'', ``Merge'' and ``Insert''.

\textit{Split}. A ``Splitting ratio'' slider is used to adjust the splitting ratio $\mu$ for vectorization. Small $\mu$ will break up long curves and generate short polylines. In contrast, large $\mu$ will retain long curves. The default value of $\mu$ is set as $160$.

\textit{Pick}. The ``Pick'' tool allows users to interactively select expressive polylines. By default, the vectorized outside OCs are automatically selected as the target as shown in Fig. \ref{fig: user_interface}(a). However, the inside polylines require manual selection because not all the simplified SCs and ARs are appropriate for stylized painting. Once a polyline is selected, the users can extend the polyline by inserting new points.

\textit{Delete}. The ``Delete'' tool is used to remove incorrectly picked polylines.

\textit{Merge}. For over-simplified or over-segmented curves, the ``Merge'' tool can be used to re-connect broken polylines and form long curves.

\textit{Insert}. The ``Insert'' tool is used to add new polylines. The missing features may happen during the feature extraction step or after neural simplification. To ensure the expressive features are correctly added or selected, our interface refers not only to the vectorized contours but also to the original model as shown in Fig. \ref{fig: user_interface}(b).

Moreover, we design two more sliders to control the maxima and minima width (thickness) of a stroke. While the interaction process is not fully automatic, users can obtain expressive strokes according to their aesthetic understandings by taking advantage of the editing tools.

\section{STROKE OPTIMIZATION AND MAPPING}  \label{sec: stroke_optim}
Unlike digital ink paintings, it is difficult to precisely control and draw physical strokes by using a robotic arm with a soft brush. Thus, we optimize the stroke trajectory and map the stroke positions and thicknesses from simulation space to physical space instead of directly input them to robotic arm.

\begin{figure*}[th]
	\centering
	\includegraphics[width=\textwidth]{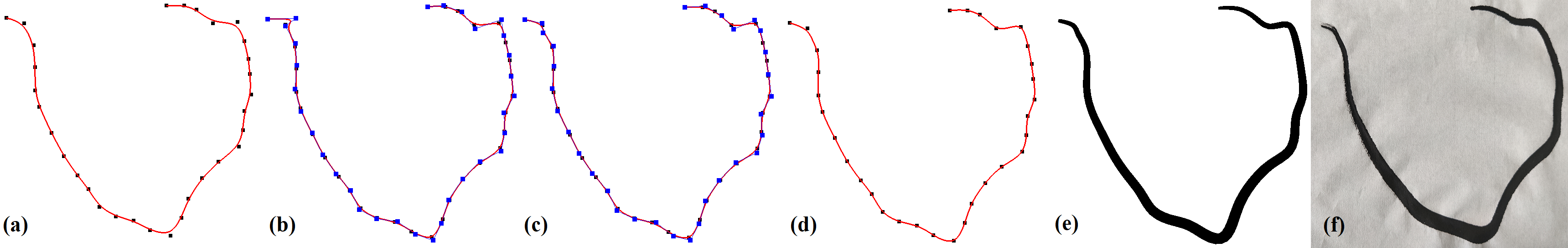}
	\captionsetup{justification=raggedright}
	\caption{Stroke optimization for robotic drawing. (a) The B-spline curve generated without optimization; The fitting curve generated by our  method in (b) 1 iteration, (c) 5 iterations, and (d) 15 iterations; (e) The simulated stroke; (f) The robotic drawing stroke with $\lambda= 1/2$ and $T_{dip}=3.0sec$. (Blue dots are control points, and black dots are sampling points.)}
	\label{fig:stroke_optim}
\end{figure*}

\subsection{Stroke Optimization}
The extracted polylines may not be smooth or unevenly sampled, which are not suitable for direct use in robotic drawing. A B-spline curve can generate evenly distributed smooth curve with several control points and avoid the Runge phenomenon. However, the choosing of control points is crucial because the generated curve does not necessarily pass through the sample points as shown in Fig. \ref{fig:stroke_optim}(a). Considering a cubic B-spline curve $C(t) = \sum_{i=1}^{m} N_{i,3}(t)P_i$, where $P_i$ is control point and $N_{i,3}(t)$ is the basis B-spline of degree $3$. To compute a smooth curve for approximating endpoints $X_k$ of extracted lines, we employ a nonlinear optimization method \cite{Wang2006} to fit $X_k$ using the following object function:
\begin{equation}\label{eqn_bspline}
	\begin{aligned}
		f = \frac{1}{2} \sum_{k=1}^{n} D_{sdm}^2(C(t) - X_k) + \alpha f_1 + \beta f_2
	\end{aligned}
\end{equation}
where $D_{sdm}(C(t) - X_k)$ is the squared distance error term to measure the distance between the point $X_k$ and the curve $C(t)$, $f_1 = \int||C'(t)||^2dt$ and $f_2 = \int||C''(t)||^2dt$ are smoothness terms, and $\alpha,\beta \geq 0$ are constant coefficients. The minimization problem of the object function $f$ can be solved by the Quasi-Newton iteration method to generate an updated fitting curve as shown in Fig. \ref{fig:stroke_optim}(b-d). After 15 iterations, the updated curve almost pass through all the sample points as seen in Fig. \ref{fig:stroke_optim}(d). 

\subsection{Thickness Setting}
The thickness of a stroke $t_i \left\lbrace i=1,2,...,n\right\rbrace$ at each of sampling vertices $C_i(x_i, y_i)$ on the optimized B-spline curve is defined by an exponential interpolation of a minimum thickness $t_{min}$ and a maximum thickness $t_{max}$ in the middle of the curve:
\begin{equation}\label{eqn_thick}
	\begin{aligned}
		t_i=
		\begin{cases}
			[1-(\frac{2i}{n})^\gamma]t_{min}+(\frac{2i}{n})^\gamma t_{max}& \text {$i<\frac{n}{2}$} \\
			[1-(\frac{2(n-i)}{n})^\gamma]t_{min}+(\frac{2(n-i)}{n})^\gamma t_{max}& \text {$i \geq \frac{n}{2}$}
		\end{cases}
	\end{aligned}
\end{equation}
where $\gamma=0.6$ is an exponent to control the change rate of the stroke thickness \cite{Grabli2010}. Fig. \ref{fig:stroke_optim}(e) shows the simulated drawing stroke with thickness controlled by Eq. (\ref{eqn_thick}) when the default values of $t_{max}$ and $t_{min}$ were set as $25$ and $4$ respectively.

\begin{table}[!h]
	\centering
	\caption{Stroke mapping to physical space}
	\label{tab:stroke_mapping}
		\begin{tabular}{m{1.2cm}<{\centering}|m{0.06cm}<{\centering}m{0.06cm}<{\centering}m{0.05cm}<{\centering}m{0.07cm}<{\centering}m{0.25cm}<{\centering}m{0.25cm}<{\centering}m{0.28cm}<{\centering}m{0.31cm}<{\centering}m{0.31cm}<{\centering}}
			\hline
			$h_i-h_{tip}$&2&4&6&8&10&12&14&16&18 \\
			\hline
			$t_i (mm)$&3&4&5&6.5&10.5&12&13&14&14.5 \\
			\hline
			&& \\
			Robotic drawing strokes&$\includegraphics[scale=0.25]{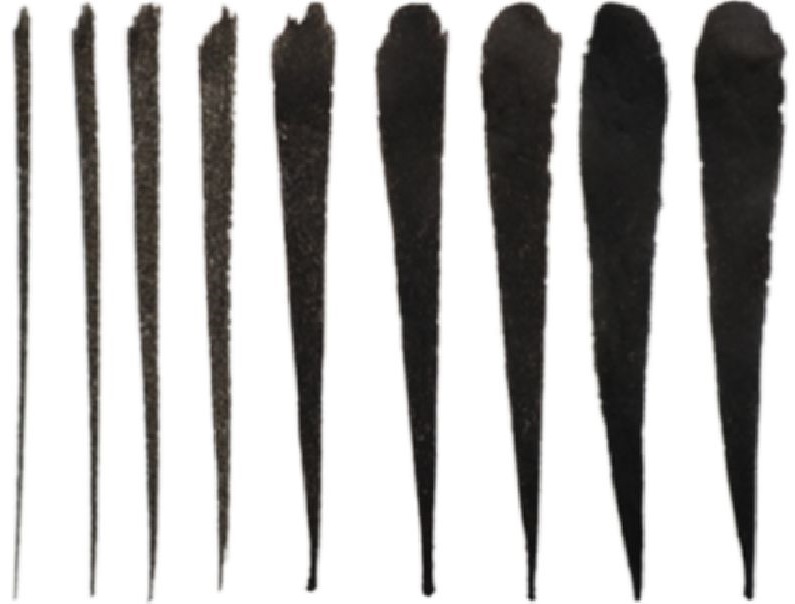}$ \\
			&&&&&&&&&\\
			\hline
	\end{tabular}
\end{table}
	
\subsection{Stroke Mapping}
In order to physically draw strokes that are consistent to the simulated strokes as shown in Fig. \ref{fig:stroke_optim}(e), firstly, we carried out an experiment to derive the relationship between stroke thickness $t_i$ and the descent of brush $h_i$ which is controlled by a robotic arm. Let $h_{tip}$ be the height at which the brush tip just reaches the paper without distortion. The descent step of the brush was set as $2mm$, and $(h_i-h_{tip})$ was updated from $2mm$ to $18mm$. The footprint of the brush is like a droplet and the increasing descent results in more severe brush bending, which in turn leads to wider stroke as shown in Tab. \ref{tab:stroke_mapping}. The thickness of each stroke $t_i$ was measured at the widest part of the stroke. Then, a standard least-squares fitting approach was applied to estimate the relationship between $t_i$ and $h_i$. Finally, we obtained $h_i=w t_i+b+h_{tip}$, where $w=1.178$, $b=-0.801$, and the coefficient of determination $R^{2}$ is 0.957, which indicates that the linear fitting is appropriate for thickness prediction. Correspondingly, the stroke position $(x_i', y_i')$ for robotic drawing can be mapped by $(x_i' = w(y_i-y_c)+x_o', y_i' = w(x_i-x_c)+y_o')$, where $(x_c, y_c)$ is the centroid of sampling vertices in simulation space, and $(x_o', y_o')$ is the center of robot Cartesian space. Fig. \ref{fig:stroke_optim}(f) shows the robotic drawing stroke which is close to the simulated one after using stroke mapping.

\section{ROBOTIC DRAWING OF STYLIZED INK PAINTING} \label{sec: robot_draw}
The oriental ink paintings utilize simple brush, black ink and water to generate strokes with varying painting styles, then to depict complex scenery such as bamboos, fish, shrimps and mountains. However, it is extremely difficult for common people to master the painting styles. Therefore, we employ a robotic arm to realize typical ink painting styles, in the vision that even an amateur can draw a fair ink painting by taking advantage of the robotic arm.

\begin{figure}[h!]
	\centering
	\includegraphics[width=\linewidth]{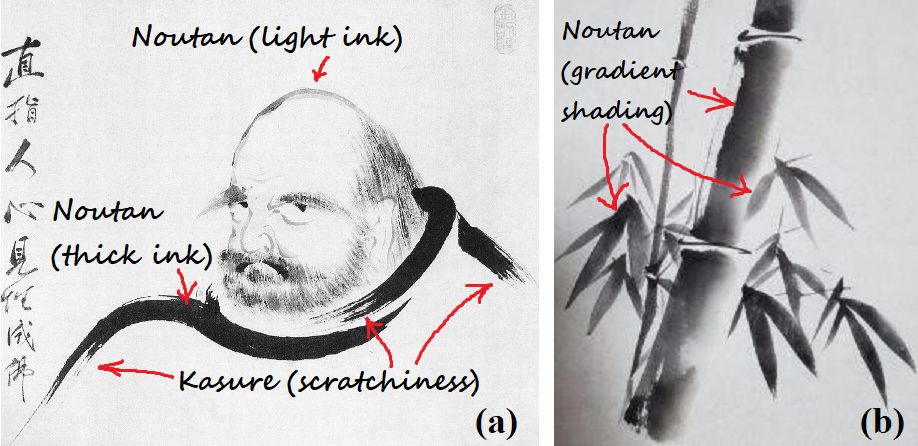}
	\captionsetup{justification=raggedright}
	\caption{Two types of painting styles in the real oriental ink paintings.(a) \textit{Kasure} (scratchiness) and \textit{Noutan} (light or thick ink); (b) \textit{Noutan} (gradient shading).}
	\label{fig:style_example}
\end{figure}

\subsection{Oriental Ink Painting Styles}
In traditional oriental ink paintings, the patterns of light and darkness can be applied to strokes by using two typical ink painting styles, which are termed as \textit{Noutan} (shade) and \textit{Kasure} (scratchiness) \cite{Zhang1999,Strassmann1986} as shown in Fig. \ref{fig: overview} and Fig. \ref{fig:style_example}. \textit{Noutan} uses water to lighten the ink and creates shade of gray color of a stroke. Fig. \ref{fig:style_example} (a, b) shows the light gray, dark gray and gradient shading strokes drawn by different \textit{Noutan} shading styles. \textit{Kasure} is the scratchy break-up of a stroke trajectory caused by insufficient and uneven supply of ink along the brush as shown in Fig. \ref{fig:style_example}(a). In our work, we explore an automatic approach to achieve the two painting styles using robotic arm.

\begin{figure}[h!]
	\centering
	\includegraphics[width=0.8\linewidth]{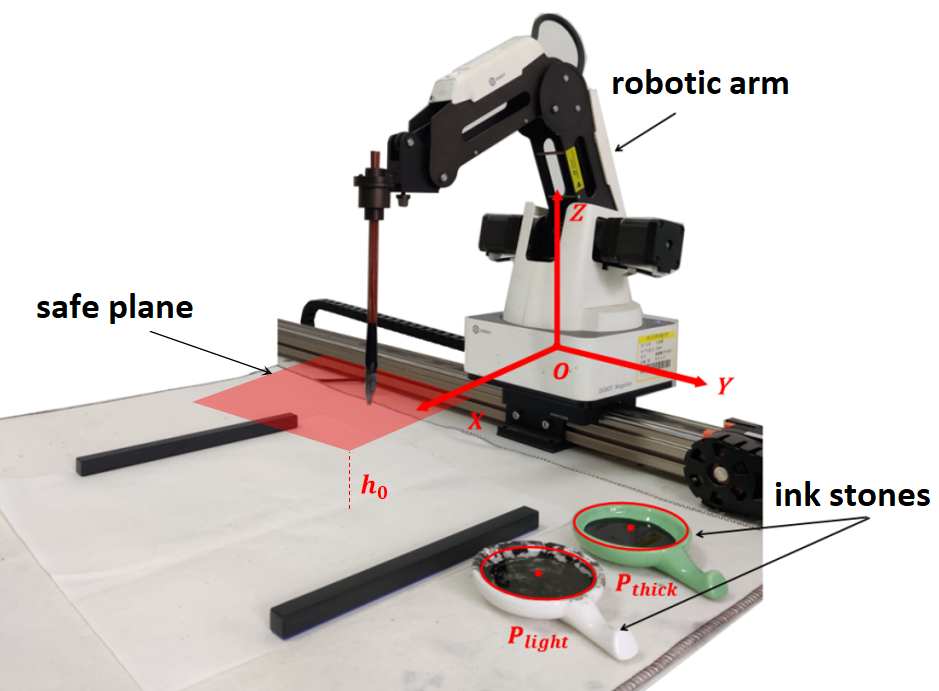}
	\captionsetup{justification=raggedright}
	\caption{Workspace of our robotic drawing framework.}
	\label{fig: robot_workspace}
\end{figure}

\subsection{Robotic Drawing of Ink Painting Styles}
To create the shading effect, we assume that \textit{Noutan} is a diffusion process of a brush with varying color of ink. Therefore, two ink stones, which are filled with light ink and thick ink, are prepared for mixing the two kinds of inks using the brush as shown in Fig. \ref{fig: robot_workspace}. 

\begin{figure}[h]
	\centering
	\includegraphics[width=\linewidth]{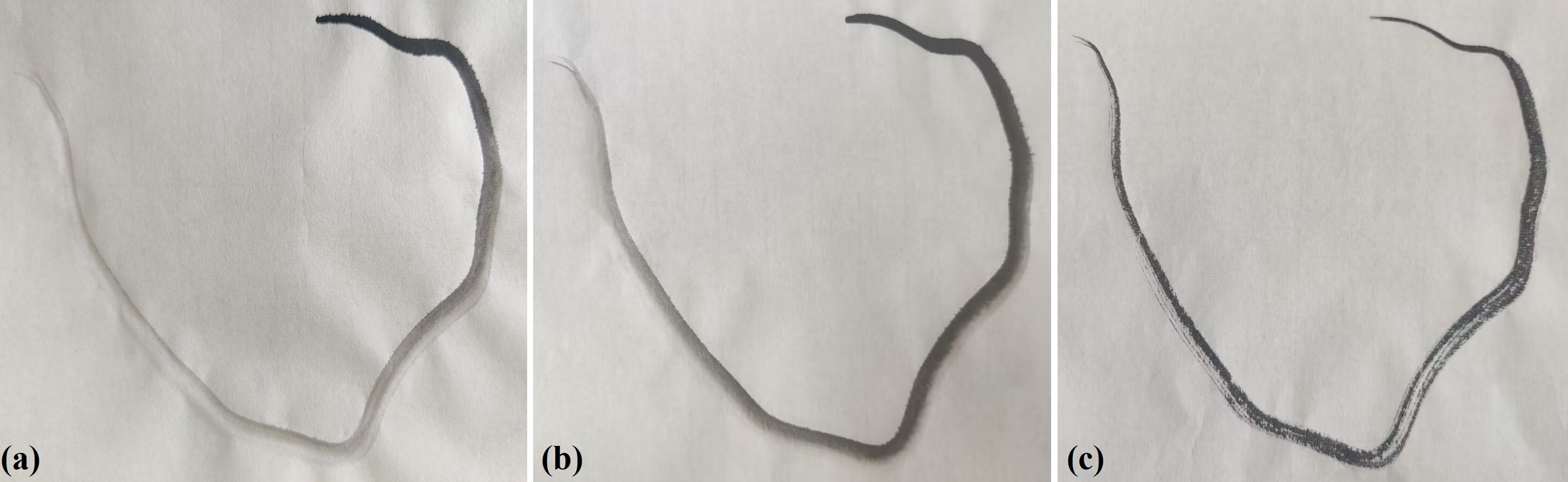}
	\captionsetup{justification=raggedright}
	\caption{Robotic drawing results of (a, b) \textit{Noutan} and (c) \textit{Kasure} styles. (a) $\lambda= 1/4$ and $T_{dip}=0.5sec$; (b) $\lambda= 1/3$ and $T_{dip}=2.5sec$; (c) $\lambda= 1/3$ and $T_{dip}=1.0sec$.}
	\label{fig: noutan_kasure}
\end{figure}

In our experiment, we observed that fully moistening the brush in the thin ink and scraping the ink at the edge of ink stone in advance is helpful for forming natural \textit{Noutan} style. Thus, we defined a series of actions for automatically rendering the \textit{Noutan} style:
\begin{equation}\label{eqn: actionNoutan}
	\begin{aligned}
		&A_{noutan}=\cup_{n=1}^{4}\{A_{tran}(P_{light}), A_{dip},\\ 
		&A_{tran}(P_{light}+((-1)^{n}r\cos(\frac{\pi}{4}), (-1)^{\lfloor n/2 \rfloor}r\sin(\frac{\pi}{4}))),\\  
		&A_{scrape}\}\cup\{ A_{tran}(P_{thick}), A_{dip}\}
	\end{aligned}
\end{equation}
where $A_{tran}$, $A_{dip}$, and $A_{scrape}$ represent the actions of translation, dipping and scraping of the brush; $P_{light}$, $P_{thick}$ and $r$ are the central positions and radii of the ink stones. Eq. (\ref{eqn: actionNoutan}) describes that the action \textit{Noutan} consists of (1) a cross-like translation, dipping and scraping in $4$ directions for fully moistening and adjusting the brush in the light ink, and (2) a translation and dipping for mixing in the thick ink. Next, the dipping time $T_{dip}$ and dipping height $L_{dip}$ of the brush in the thick ink can be applied to control the shading degree of \textit{Noutan}:
\begin{equation}\label{eqn: degreeNoutan}
	\begin{aligned}
		D_{noutan} = \frac{L_{dip}(1-e^{-c_1 T_{dip}})} {L_{brush}}
	\end{aligned}
\end{equation}
where $L_{brush}$ is the length of a brush, and $c_1$ is a constant coefficient. Let $\lambda=L_{dip}/L_{brush}$ which denotes the dipped ratio of a brush. Then, a small $\lambda$ and a small $T_{dip}$ will create bright shading effect as shown in Fig. \ref{fig: noutan_kasure}(a) because of shallow mixing in the thick ink, and a large $\lambda$ and a large $T_{dip}$ will generate dark shading effect as shown in Fig. \ref{fig: noutan_kasure}(b) because of sufficient mixing in the thick ink.

\textit{Kasure} mainly depends on the property of brush hairs, the quantity of ink absorbed by the brush and the area of drawing strokes. It is difficult to dynamically control the property of each brush hair such as the length, stiffness and the ability of absorbing fluid. However, it is possible to control the amount of ink on the brush and the area of strokes. Therefore, we assume that the degree of scratchiness is proportional to the stroke area which consists of a series of trapezoids and inversely proportional to the absorbed quantity of ink which is related to the dipping time $T_{dip}$ and the dipped ratio $\lambda$:
\begin{equation} \label{eqn_degreeKasure}
	\begin{aligned}
		D_{kasure} = \frac{c_2 \sum_{i=1}^{n-1} [(t_i+t_{i+1})||C_i-C_{i+1}||]} { 2\lambda (1-e^{-c_3 T_{dip}})}
	\end{aligned}
\end{equation}
where $c_2$ and $c_3$ are constant coefficients. In our experiment, the brush remained dry before being dipped into an ink stone. Fig. \ref{fig: noutan_kasure}(c) shows the distinguishable effect of scratchiness when setting a small $\lambda$ and a small $T_{dip}$. In contrast, the \textit{Kasure} effect only happened at the very end of the stroke when setting a large $\lambda$ and a large $T_{dip}$ as shown in Fig. \ref{fig:stroke_optim}(f).

\subsection{Trajectory Planning}
In order to draw stylized strokes using robotic arms, we first mapped the sampling positions and thicknesses $(C_i, t_i)(i=1,2,...,n)$ of an optimized stroke in simulation space to $(C'_i, h_i)$ in robot Cartesian space as introduced in Sec. \ref{sec: stroke_optim}. Then, we set the painting styles, dipping time $T_{dip}$ and dipped ratio $\lambda$ of a brush for stylized drawing. Next, the brush was lifted to an initial safe plane with a height $h_0$ above a drawing plane waiting for real executions as shown in Fig. \ref{fig: robot_workspace}. After that, a series of actions such as $A_{tran}$, $A_{dip}$ and $A_{scrape}$ were combined to form various painting styles. Finally, the stroke trajectory $(C'_i, h_i)$ integrated with the way-points of these actions were automatically converted to joints' rotations from Cartesian space to joint space by solving an inverse kinematics problem \cite{Hock2017}. One of an important design of trajectory planning is to choose a joint-space or task-space trajectory. In our framework, we used task-space trajectory since it generates physical strokes more faithful to the simulated ones. Furthermore, the speed of robotic drawing for each stroke was constrained by a typical trapezoidal velocity model which ensures piecewise trajectories of constant acceleration, zero acceleration, and constant deceleration.

\section{RESULTS AND DISCUSSION} \label{sec:results}
Our robotic drawing framework was implemented in a PC configured with a $4.0 GHz$ CPU, $16 GB$ RAM, and Windows $10$ OS. The user interface, optimization and control algorithms were developed by Python $3.6$ and C++. In order to satisfy the requirements of painting for common people, a consumer-grade robotic arm ``Dobot Magician'' (less than $\$2000$) with $0.2mm$ positioning repeatability is used to draw different painting styles. All motions of the robotic arm are realized in $4$ DOFs, and the drawing speed of the robotic arm is set from $50mm/sec$ to $120mm/sec$ for stable painting. 

\begin{figure}[h]
	\centering
	\includegraphics[width=\linewidth]{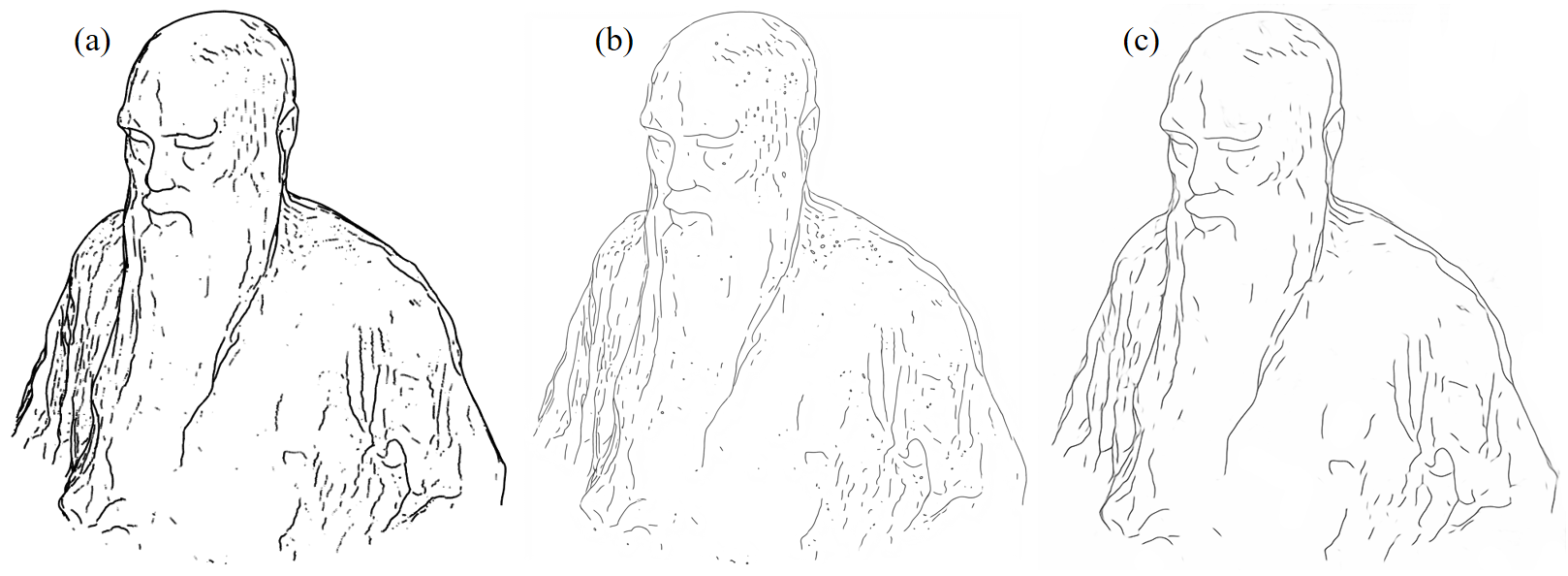}
	\captionsetup{justification=raggedright}
	\caption{Simplified contours generated from pre-trained model and updated model with augmented datasets. (a) The input rough contours of a 3D model; (b) Simplification result based on the pre-trained  model of \cite{Simo2016}; (c) Simplification result based on the updated model trained by our dataset.}
	\label{fig: network_ablation}
\end{figure}

\subsection{Evaluation of Simplification Network} Fig. \ref{fig: network_ablation} shows two different simplified contours generated from the input rough contours of a 3D model. We can observe that our updated model with augmented dataset generated cleaner and longer contours  than the results of the pre-trained model of Simo-Serra et al. \cite{Simo2016}.

\begin{figure}[h]
	\centering
	\includegraphics[width=\linewidth]{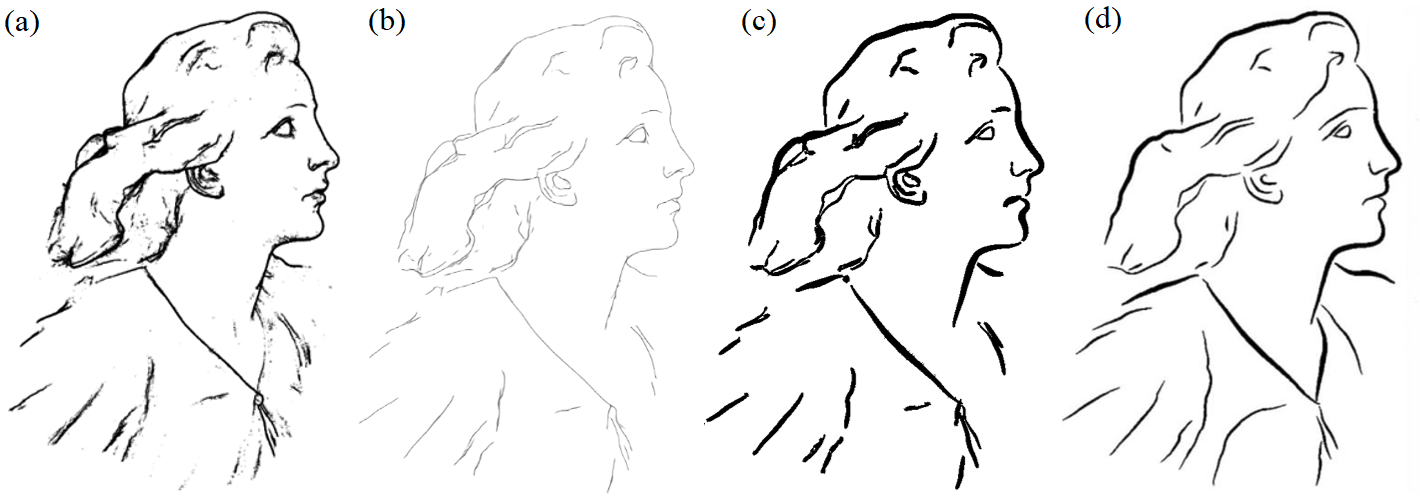}
	\captionsetup{justification=raggedright}
	\caption{Automatically generated strokes vs. interactive editing strokes from simplified contours. (a) The input rough contours of a 3D model; (b) Simplified contours; (c) Automatically generated strokes from (b); (d) Interactive editing strokes from (b).}
	\label{fig:interactive_editing}
\end{figure}

\subsection{Evaluation of Editing Tools} If we directly use all the simplified contours to generate strokes without interactive editing, it will generate many redundant strokes overlapping with each other or some expressive strokes might be ignored as shown in Fig. \ref{fig:interactive_editing}(c). Therefore, we designed the interactive editing tools to choose and modify expressive strokes by referring to both the simplified contours and the original 3D models. Fig. \ref{fig:interactive_editing}(d) shows that our editing tools can generate clean and expressive strokes.

\begin{figure}[h]
	\centering
	\includegraphics[width=\linewidth]{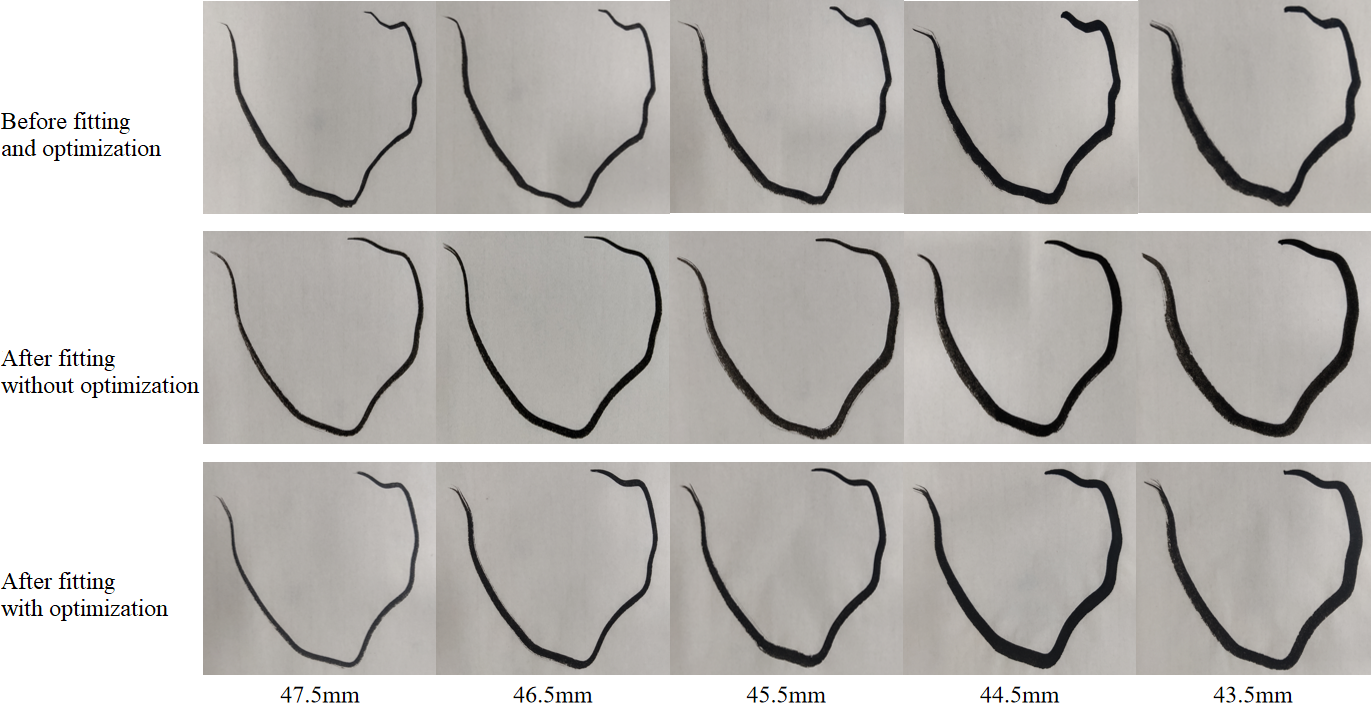}
	\captionsetup{justification=raggedright}
	\caption{Comparison of stroke drawing results with various descent of brush height $h_i$.}
	\label{fig:stroke_compare}
\end{figure}

\subsection{Evaluation of Stroke Optimization} To evaluate the effectiveness of stroke optimization algorithm, we have conducted an experiment to verify that the physically drawing stroke can keep the shape of a simulated curve when the brush is deformed with various descent height $h_i$ as shown in Fig. \ref{fig:stroke_compare}. The first row of Fig. \ref{fig:stroke_compare} show the robotic drawing results of strokes before B-spline fitting and optimization. The second row of Fig. \ref{fig:stroke_compare} shows the robotic drawing results of strokes after B-spline fitting without optimization, and the third row of Fig. \ref{fig:stroke_compare} shows the robotic drawing results of strokes after B-spline fitting with optimization.

\begin{figure}[h]
	\centering
	\includegraphics[width=0.4\linewidth]{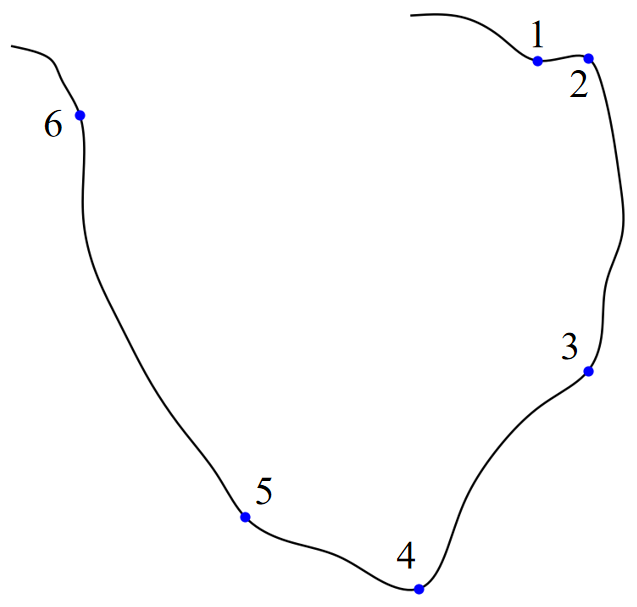}
	\captionsetup{justification=raggedright}
	\caption{The typical $6$ turning points of a target curve.}
	\label{fig:turning_points}
\end{figure}

\begin{figure*}[bp]
	\centering
	\includegraphics[width=\textwidth]{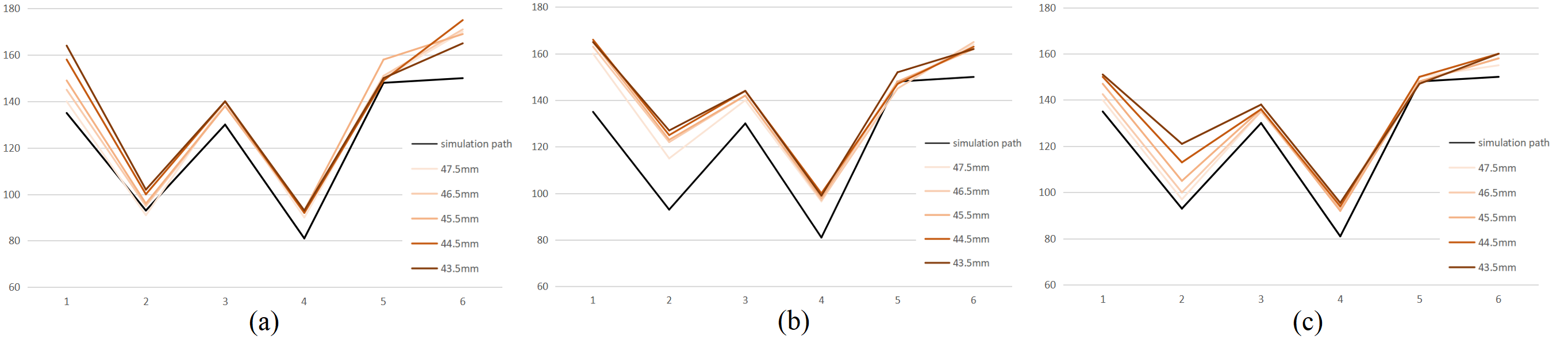}
	\captionsetup{justification=raggedright}
	\caption{Changes of the $\theta$ at the $6$ turning points (a) before fitting and optimization, (b) after fitting without optimization and (c) after fitting with optimization.}
	\label{fig:angle_deformation}
\end{figure*}

To quantitatively evaluate the stroke optimization method, we utilized the angle of contingence $\theta$ (or the slope change between contiguous straight-line segments at a point) presented by \cite{Bribiesca2013} to represent the discrete curvature at the $6$ typical turning points of the simulated curve as shown in Fig. \ref{fig:turning_points}. The original angles of contingence of the turning points in the simulated curve are $135^{\circ}$, $93^{\circ}$, $130^{\circ}$, $81^{\circ}$, $148^{\circ}$ and $150^{\circ}$. The angles of contingence corresponding to the three kinds of physically drawing strokes in Fig. \ref{fig:stroke_compare} are shown in Tab. \ref{tab:angle_deformation_1}, \ref{tab:angle_deformation_2} and \ref{tab:angle_deformation_3}. Fig. \ref{fig:angle_deformation}(a-c) show the corresponding changing trends of $\theta$ from Tab. \ref{tab:angle_deformation_1}, \ref{tab:angle_deformation_2} and \ref{tab:angle_deformation_3}.

\begin{table}[h]
	\caption{Statistics of $\theta$ before fitting and optimization}
	\centering
	\label{tab:angle_deformation_1}
	\begin{tabular}{c|ccccccm{0.8cm}<{\centering}}
		\hline
		descent height& 1 & 2 & 3 & 4 & 5 & 6 & average error\\
		\hline
		47.5mm& 140 & 91 & 140 & 90 & 150 & 170 &  \textbf{7.33}\\
		46.5mm& 145 & 95 & 138 & 92 & 151 & 171 & \textbf{9.17}\\
		45.5mm& 149 & 96 & 138 & 93 & 158 & 169 & \textbf{11.00}\\
		44.5mm& 158 & 100 & 140 & 92 & 149 & 175 & \textbf{12.83}\\
		43.5mm& 164 & 102 & 140 & 93 & 150 & 165 & \textbf{12.83}\\
		\hline
	\end{tabular}
\end{table}

\begin{table}[h]
	\caption{Statistics of $\theta$ after fitting without optimization}
	\centering
	\label{tab:angle_deformation_2}
	\begin{tabular}{c|ccccccm{0.8cm}<{\centering}}
		\hline
		descent height& 1 & 2 & 3 & 4 & 5 & 6 & average error\\
		\hline
		47.5mm& 160 & 115 & 140 & 96.5 & 147 & 164 & \textbf{14.25}\\
		46.5mm& 163 & 122 & 142 & 97 & 145 & 165 & \textbf{16.17}\\
		45.5mm& 165 & 123 & 142 & 98 & 148 & 169 & \textbf{16.83}\\
		44.5mm& 166 & 125 & 144 & 100 & 147 & 175 & \textbf{18.00}\\
		43.5mm& 165 & 127 & 144 & 99 & 152 & 165 & \textbf{18.67}\\
		\hline
	\end{tabular}
\end{table}

\begin{table}[h]
	\caption{Statistics of $\theta$ after fitting with optimization}
	\centering
	\label{tab:angle_deformation_3}
	\begin{tabular}{c|ccccccm{0.8cm}<{\centering}}
		\hline
		descent height& 1 & 2 & 3 & 4 & 5 & 6 & average error\\
		\hline
		47.5mm& 140 & 97 & 135 & 95 & 150 & 155 &  \textbf{5.83}\\
		46.5mm& 142.5 & 100 & 135 & 93 & 148 & 158 & \textbf{6.58}\\
		45.5mm& 147 & 105 & 136 & 92 & 148 & 158 & \textbf{8.16}\\
		44.5mm& 150 & 113 & 136 & 94 & 150 & 160 & \textbf{11.00}\\
		43.5mm& 151 & 121 & 138 & 95.5 & 147 & 160 & \textbf{12.58}\\
		\hline
	\end{tabular}
\end{table}

From Tab. \ref{tab:angle_deformation_1} and \ref{tab:angle_deformation_2}, we can observe that the total average error of $\theta$ after fitting is $16.784$, that is much larger than that of angle of contingence ($10.632$) before fitting, though the fitted curve generated more smooth strokes as shown in Fig. \ref{fig:stroke_compare}. From Tab. \ref{tab:angle_deformation_1} and \ref{tab:angle_deformation_3}, we can conclude that the optimized B-spline curve not only create the minimal total average error $8.83$, but also generate smooth strokes. Therefore, the proposed B-spline fitting and optimization method has the merits of retaining the shape of the simulated stroke as well as generating physically smooth strokes.

\begin{figure}[h]
	\centering
	\includegraphics[width=\linewidth]{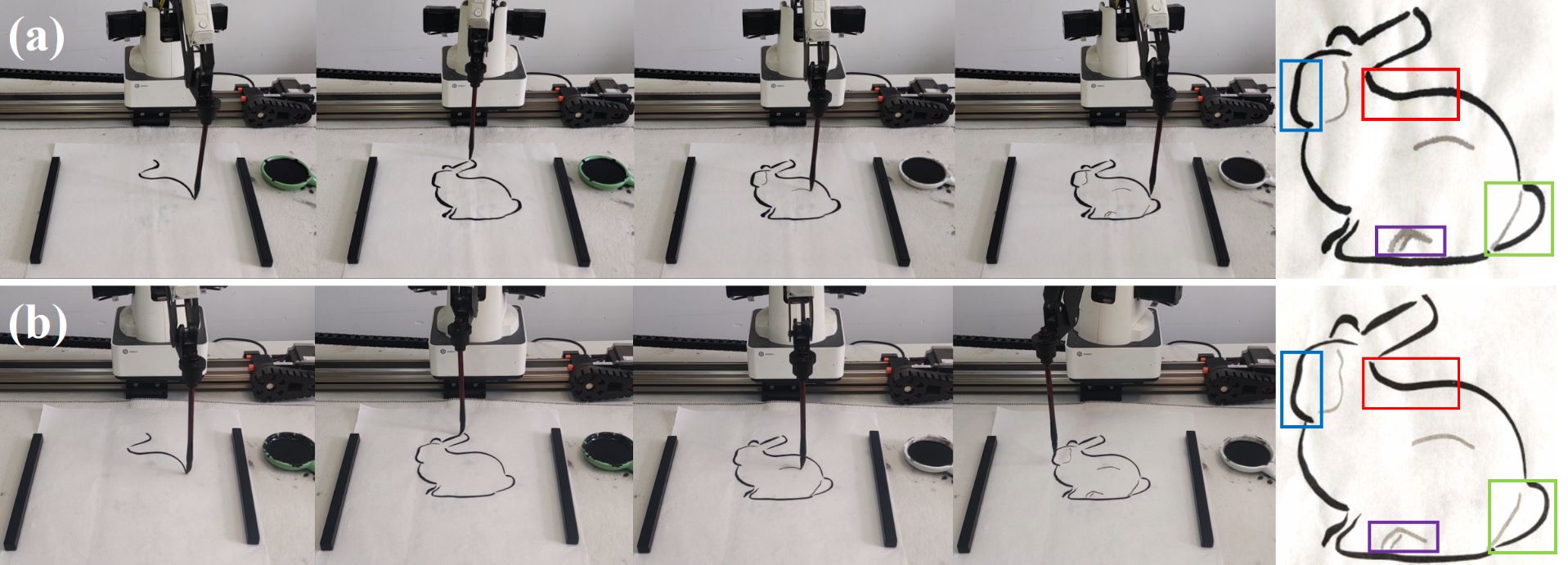}
	\captionsetup{justification=raggedright}
	\caption{Robotic drawing process of ``Stanford Bunny'' (a) before and (b) after applying stroke optimization.}
	\label{fig: compare_optim}
\end{figure}

\begin{figure}[h]
	\centering
	\includegraphics[width=\linewidth]{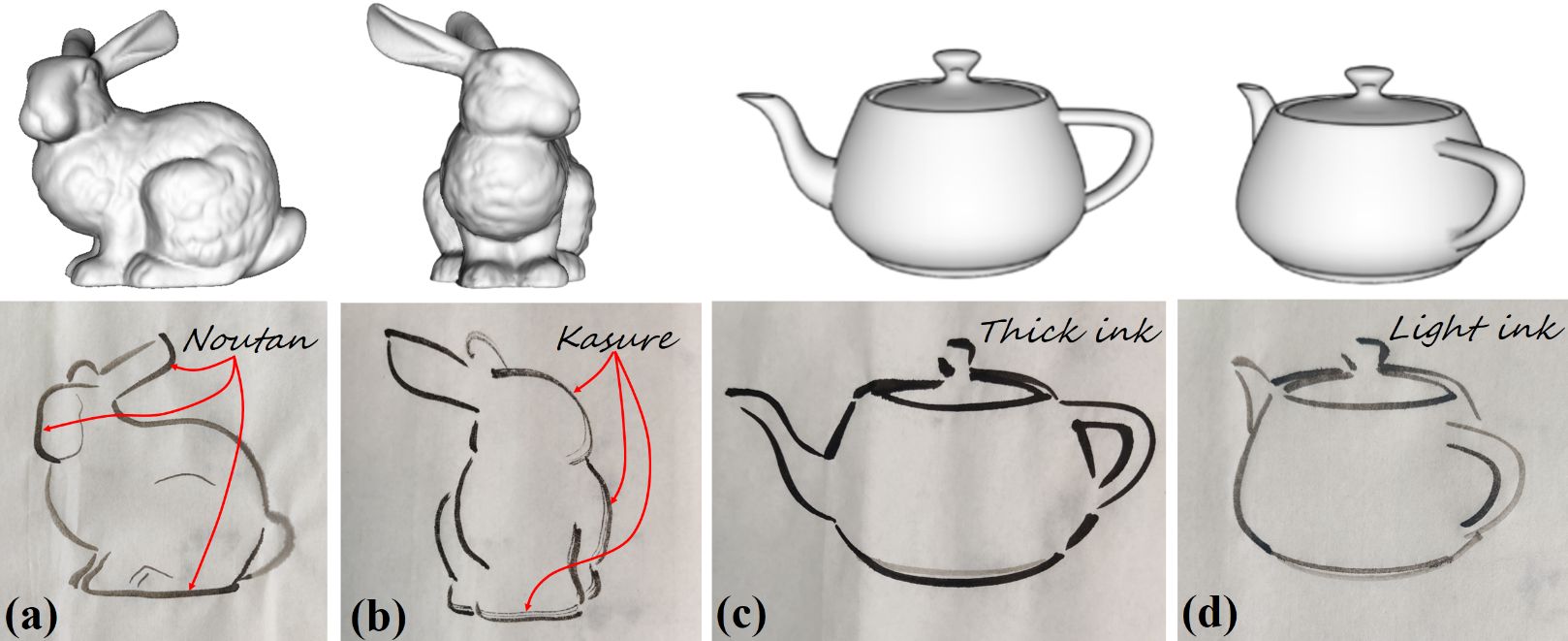}
	\captionsetup{justification=raggedright}
	\caption{Robotic drawing results with two painting styles including (a) \textit{Noutan} (gradient shading), (b) \textit{Kasure}, (c) \textit{Noutan} (thick ink), and (d) \textit{Noutan} (light ink).}
	\label{fig: ink_styles}
\end{figure}

Furthermore, we have compared the drawing results of the ``Stanford Bunny" before and after using the optimization method. It can be observed that the optimized strokes are more smooth and natural than the original strokes before optimization as shown in the colored rectangles of Fig. \ref{fig: compare_optim}.

\subsection{Evaluation of Painting Styles} To test the oriental ink painting styles, we have conducted an experiment to show the effect of \textit{Noutan} and \textit{Kasure} on ``Stanford Bunny'' and ``The Utah Teapot'' from different viewpoints. Fig. \ref{fig: ink_styles}(a) shows the gradient shading effect of \textit{Noutan} style and Fig. \ref{fig: ink_styles}(b) shows the scratchy effect of \textit{Kasure} style. The extreme case of \textit{Noutan} style is using thick ink or light ink only to convey strong or delicate impression as shown in Fig. \ref{fig: ink_styles}(c, d).

To quantitatively measure the intensity of the two ink painting styles, we set $12$ different combinations of dipped ratio $\lambda$ and dipping time $T_{dip}$, and then draw the painting with \textit{Noutan} and \textit{Kasure} styles on rice paper, and $2$ experts were invited to score each painting.

\begin{figure*}[ht]
	\centering
	\includegraphics[width=1.0\textwidth]{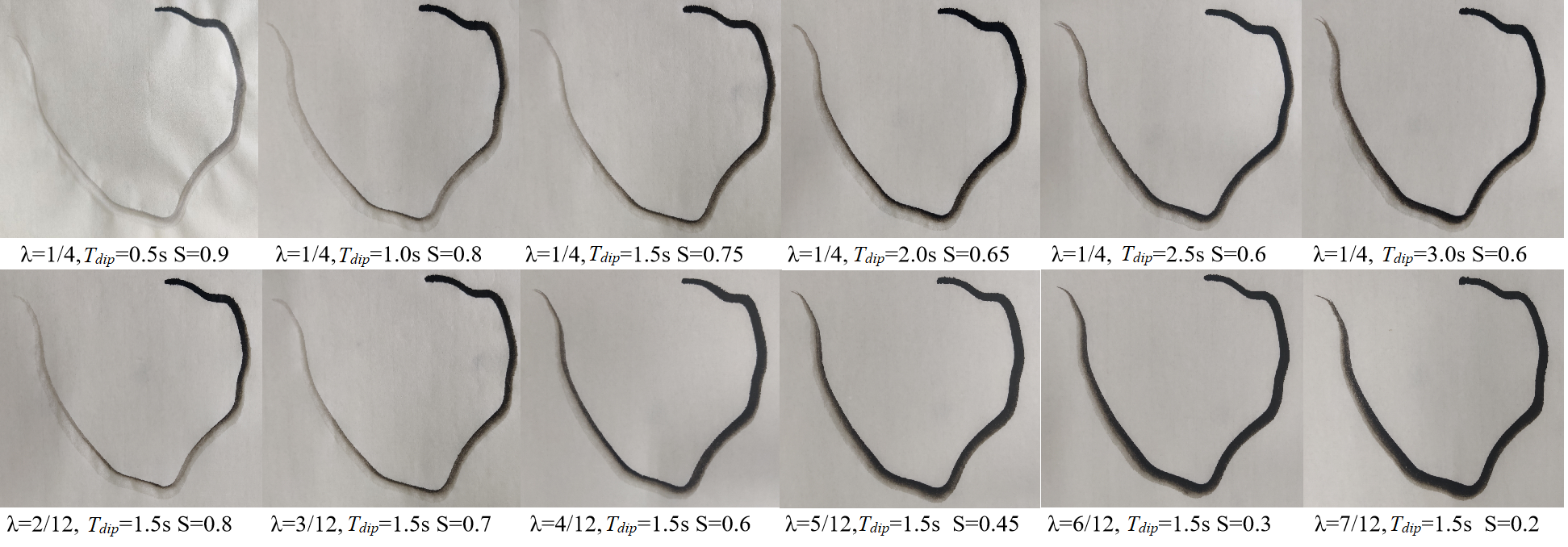}
	\captionsetup{justification=raggedright}
	\caption{Results of \textit{Noutan} effect with different combinations of $\lambda$ and $T_{dip}$.}
	\label{fig:noutan_experiment}
\end{figure*}

\begin{figure*}[bp]
	\centering
	\includegraphics[width=1.0\textwidth]{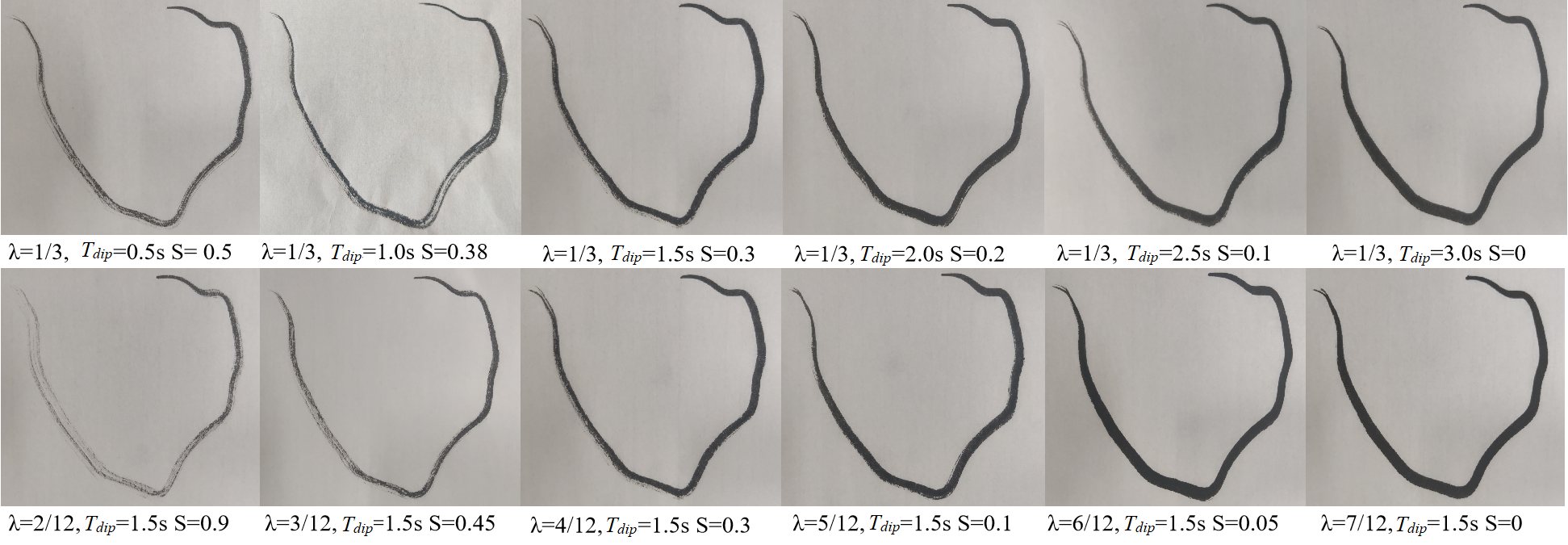}
	\captionsetup{justification=raggedright}
	\caption{Results of \textit{Kasure} effect with different combinations of $\lambda$ and $T_{dip}$.}
	\label{fig:kasure_experiment}
\end{figure*}

Fig. \ref{fig:noutan_experiment} shows the results of \textit{Noutan} experiment. To observe the different effect of parameters, we changed one parameter while keeping the remaining parameter constant. The first row of Fig. \ref{fig:noutan_experiment} shows the score of \textit{Noutan} effect when $\lambda$ is constant and $T_{dip}$ is changing, and the second row of Fig. \ref{fig:noutan_experiment} shows the score of \textit{Noutan} effect when $T_{dip}$ is constant and $\lambda$ is changing. As the dipping time is gradually increased from $0.5s$ to $3.0s$, the \textit{Noutan} intensity decreases and converges to $0.6$. On the other hand, when the dipped ratio is increased from $2/12$ to $7/12$, the \textit{Noutan} intensity gradually decreases from $0.8$ to $0.2$.

Fig. \ref{fig:kasure_experiment} shows the results of \textit{Kasure} experiment. The relationship of the dipping time and the dipped ratio are similar to the $Noutan$ experiment. Moreover, we can observe that the dipped ratio has a greater impact on style intensity than the dipping time.

\begin{figure*}[h]
	\centering
	\includegraphics[width=0.85\linewidth]{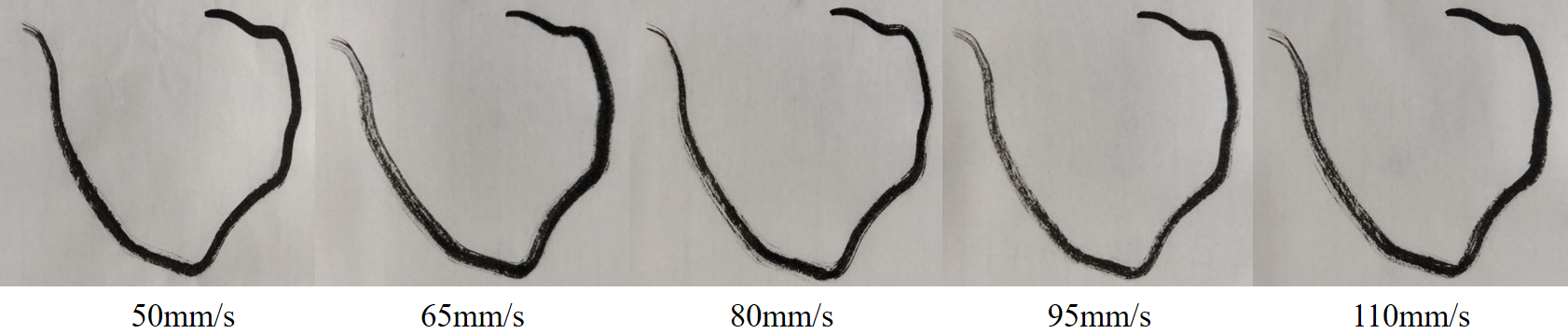}
	\captionsetup{justification=raggedright}
	\caption{Drawing results of \textit{Kasure} style of a stroke with different robot's velocity.}
	\label{fig:kasure_brush_speed_deformation}
\end{figure*}

Furthermore, we have designed an experiment to demonstrate the effect of robot's velocity on \textit{Kasure} style of a stroke and the drawing results are shown in Figure \ref{fig:kasure_brush_speed_deformation}. The experiment was conducted with the dipped ratio $\lambda=1/3$ and the dipping time $T_{dip}=2.0s$. The drawing velocity $v$ was set from $50mm/s$ to $110mm/s$ with an increasing step $15mm/s$. From the results, we can observe that the increasing of robot's velocity will result in more intensive \textit{Kasure} effect at the initial stage from $50mm/s$ to $95mm/s$. However, the \textit{Kasure} effect will converge as the speed increases at the closing stage from $95mm/s$ to $110mm/s$. The results showed that robot's velocity does impact the final painting styles.

However, there are many factors possibly impact the painting styles such as the drawing speed, the property of brush hairs, the quantity of ink absorbed by the brush and the area of drawing strokes. In our work, it is difficult to consider all these factors for controlling a specific painting style. Inspired by Strassmann's pioneer work of non-photorealistic rendering ``Hairy Brushes'', we fixed the drawing velocity and mainly consider the factor ink quantity because the ink supply on each bristle of a brush is assumed to be a reservoir of a finite quantity of fluid and the quantity is decreased as the brush moves through a stroke \cite{Strassmann1986}. Therefore, our framework chooses to generate ink style by controlling ink quantity instead of robot's velocity.

\begin{figure*}[ht]
	\centering
	\includegraphics[width=\textwidth]{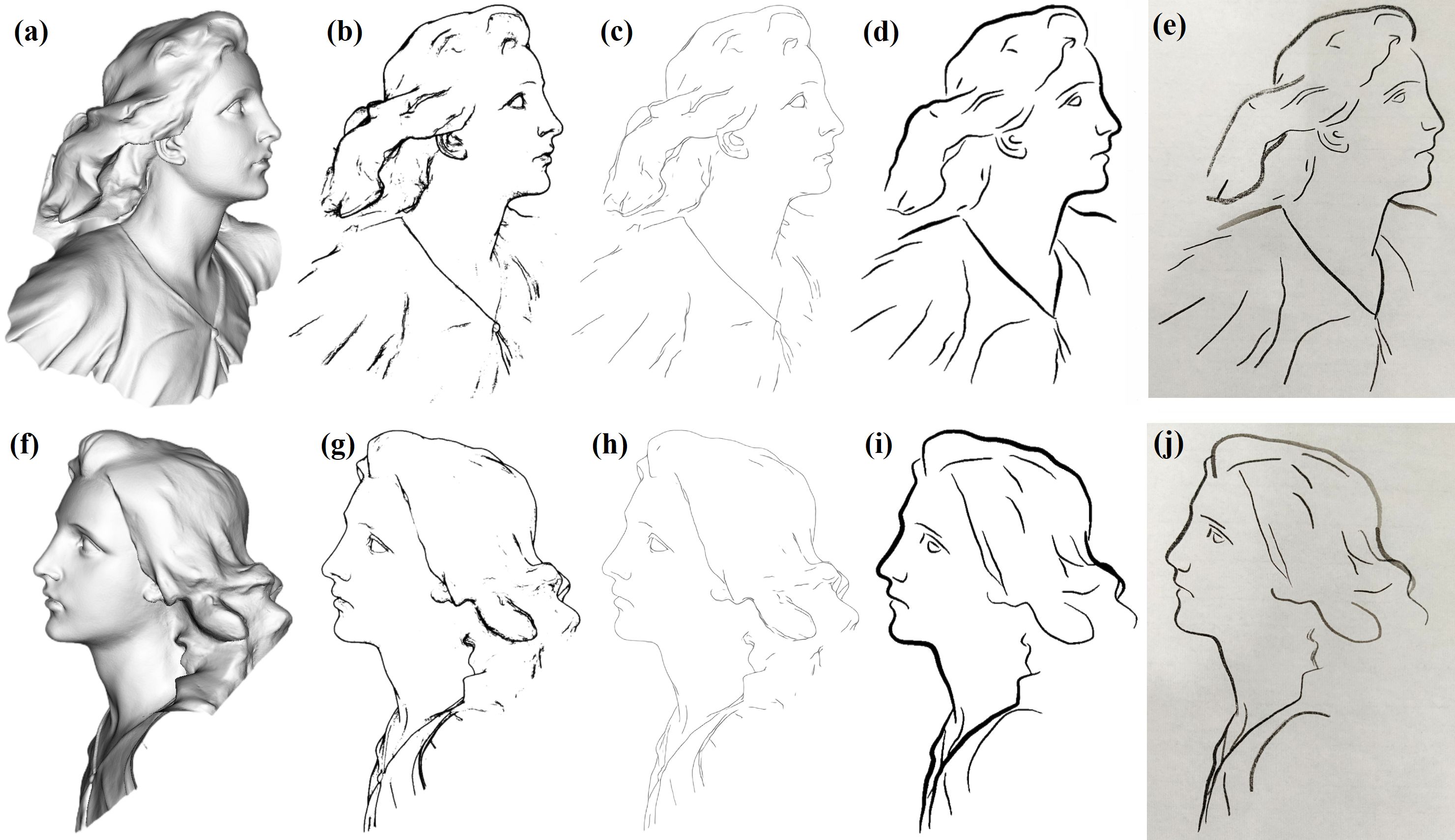}
	\captionsetup{justification=raggedright}
	\caption{The physical ink painting generation process of ``Lucy''. (a, f) Left and right views; (b, g) Initial geometrical contours; (c, h) Simplified contours; (d, i) Simulated ink paintings after vectorization and user editing; (e, j) Robotic drawing results.}
	\label{fig: compare_lucy}
\end{figure*}

\subsection{Applications} We also examined the ability of our framework to draw complicated stylized portrait from the standard 3D model ``Lucy'' as shown in Fig. \ref{fig: compare_lucy}. The results showed that the physically drawing strokes retained the characteristics of reference model after feature extraction, simplification, digital simulation and optimization. Note that the \textit{Kasure} and \textit{Noutan} styles have been applied to the hair and shoulder of ``Lucy'' respectively. 

\begin{figure*}[h]
	\centering
	\includegraphics[width=\textwidth]{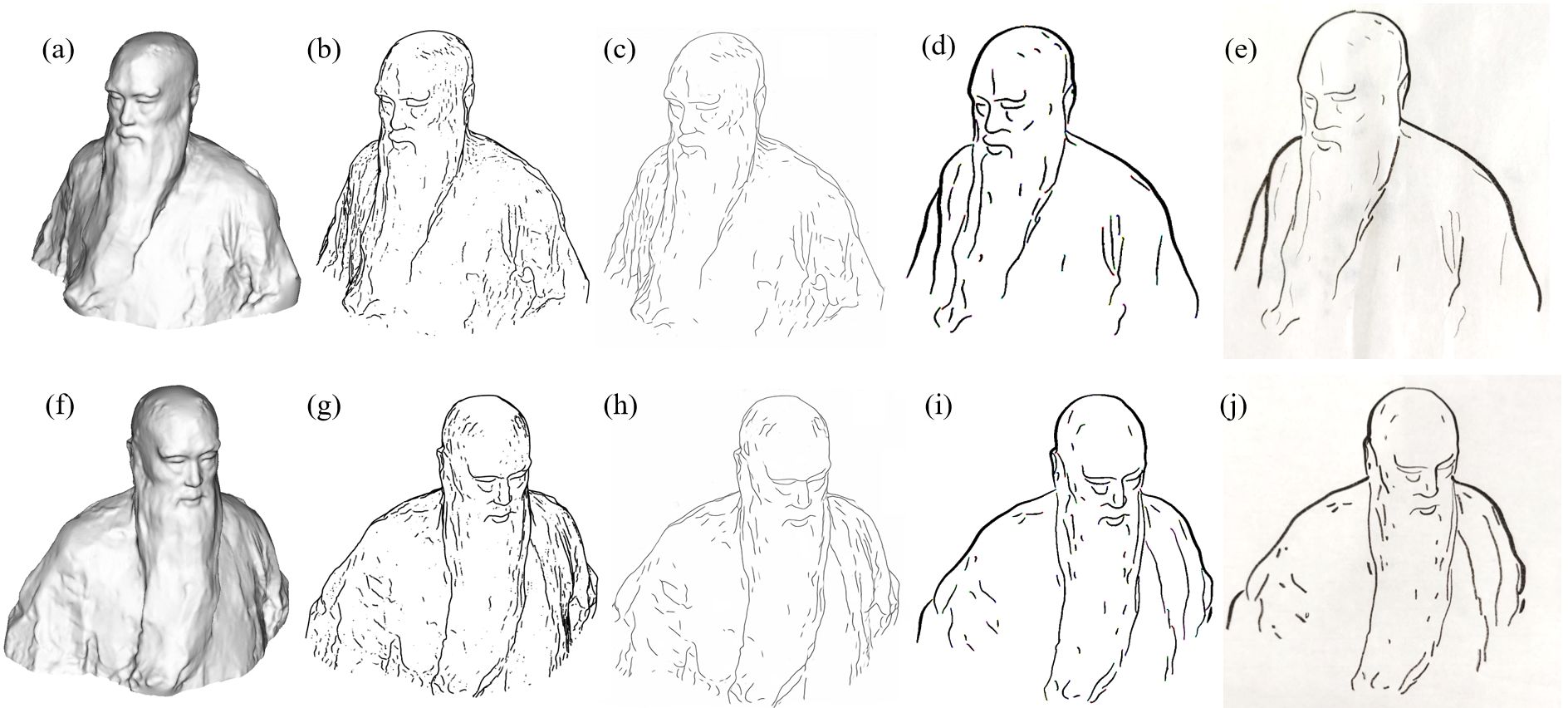}
	\captionsetup{justification=raggedright}
	\caption{Robotic drawing results of a user-defined 3D model ``Youren''. (a, f) Two different views; (b, g) Initial contours; (c, h) Simplified contours; (d, i) Simulated results; (e, j) Robotic drawing results.} 
	\label{fig: compare_youren}
\end{figure*} 

To evaluate the scalability of our robotic drawing framework for arbitrary user-defined model, we collected approximately 40 photos around a sculpture named ``Youren'' by a normal smartphone, and then reconstructed its point cloud and surface by using SfM (Structure from Motion) and screened Poisson reconstruction algorithms. Fig. \ref{fig: compare_youren} shows that a small number of stylized strokes can depict the facial expression of a character, which indicates that our framework has the potential for drawing vivid portraits from varying viewpoints once a 3D model is given.

\subsection{Performance} We calculated the time of line extraction, stroke optimization and robotic drawing of the tested $4$ 3D models, each of which has 2 viewpoints, and the quantitative results of our robotic drawing framework is shown in Tab. \ref{tab:time_statistics}.

\begin{table}[h]
	\caption{Time statistics of the robotic drawing framework.}
	\centering
	\label{tab:time_statistics}
	\begin{tabular}{m{0.6cm}<{\centering}|m{1.0cm}<{\centering}|m{1.2cm}<{\centering}|m{1.2cm}<{\centering}|m{1.0cm}<{\centering}|m{1.0cm}<{\centering}}
		\hline
		model                   & viewpoint &number of strokes &line extraction & stroke optim. & robotic drawing\\
		\hline
		\multirow{2}{*}{teapot} & V1        & 16                    & 86s             & 3.9s                & 322s              \\
		& V2        & 14                    & 78s             & 3.8s                & 316s              \\
		\hline
		\multirow{2}{*}{bunny}  & V1        & 14                    & 84s             & 3.8s                & 303s              \\
		& V2        & 12                    & 75s             & 3.6s                & 356s              \\
		\hline
		\multirow{2}{*}{Lucy}   & V1        & 44                    & 315s            & 14.7s               & 723s              \\
		& V2        & 29                    & 291s            & 8.9s                & 839s              \\
		\hline
		\multirow{2}{*}{Youren} & V1        & 51                    & 512s            & 16.3s               & 450s              \\
		& V2        & 76                    & 762s            & 24.1s               & 541s             \\
		\hline
	\end{tabular}
\end{table}

Although strokes have various length and thickness, the average execution time of line extraction, stroke optimization and robotic drawing of each stroke is $8.61s$, $0.31s$ and $15.04s$. The results showed that the robotic drawing of stroke is the most time-consuming stage because the physical drawing of a stroke with a brush is inherently slow, and the expressive line extraction is the second time-consuming stage because it requires user interaction, and the stroke optimization is the most efficient stage because it is fully automatic.

\section{Conclusion} \label{sec: concl}
We presented a new robotic drawing framework for converting 3D models to oriental ink paintings. With a combination of automatic contour extraction methods and human-computer interaction process, the framework is flexible for users to create personalized expressive digital strokes in simulation stage. Furthermore, through the proposed stroke optimization, mapping and motion planning methods, the framework is able to draw stylized ink paintings in real execution stage, which is the key contribution of our work.

However, the current framework is not fully automatic because the selection of expressive contours depends on different users and the manually annotated data is not sufficient for training a comprehensive network for arbitrary 3D models. In the future, we are going to collect and train a contour selection network from painting experts, and ultimately achieve an end-to-end framework for a more streamlined solution. Secondly, the current drawing focuses on contours instead of shadows and textures. We intend to learn the diffusion effect of ink paintings by considering advanced NPR skills such as hatching. Another restriction is that the impact of drawing speed is not considered in robotic drawing. We will improve the performance of painting styles by fine-grained velocity-intensity model. Furthermore, it is noticed that the DynamicFusion \cite{Newcombe2015} and SplitFusion \cite{Li2020} are very promising 3D vision works for generating 3D models in real-time. We plan to use these methods to generate 3D models in real-time and draw the corresponding artworks with ink painting style in our future work.

For a better understanding of the results of interactive editing, optimization and robotic drawing, please refer to the supplementary video from the submitted attachment or the link (\url{https://cie.nwsuaf.edu.cn/docs/2023-08/630b0573861443c3ae1366de3cef1263.mp4}).

\section*{Acknowledgement}
We would like to gratefully thank the associate editor and the reviewers for their constructive comments and suggestions. This work was supported in part by the Natural Science Basis Research
(NSBR) Plan of Shaanxi under Grant 2022JM-363, the Key Project of Shaanxi Provision-City Linkage under Grant 2022GD-TSLD-53 and the National Natural Science Foundation of China under Grant 61303124. 

\ifCLASSOPTIONcaptionsoff
  \newpage
\fi



\bibliographystyle{IEEEtran}
\bibliography{references_IEEE_format}
\end{document}